\documentclass[11pt]{article}
\usepackage{amsmath}
\usepackage{amssymb}
\usepackage{enumitem}
\usepackage{tabularx}
\usepackage{placeins}
\usepackage{algorithm}
\usepackage{algpseudocode}
\usepackage[final]{acl}

\usepackage{times}
\usepackage{latexsym}
\usepackage{booktabs}
\usepackage{makecell}
\usepackage{pifont}
\usepackage{multirow}
\usepackage{xcolor}
\usepackage{listings}
\usepackage{tcolorbox}
\tcbuselibrary{skins,breakable}
\usepackage[T1]{fontenc}

\usepackage[utf8]{inputenc}

\usepackage{microtype}

\usepackage{inconsolata}

\usepackage{graphicx}
\usepackage{subcaption}

\title{Shopping Companion: Benchmarking and Training LLM Agents for Long-Horizon Preference-Grounded E-Commerce Tasks}

\author{Zijian Yu$^{*}$ \quad Kejun Xiao$^{*\dagger}$ \quad Huaipeng Zhao \quad Tao Luo \quad Xiaoyi Zeng \\
        Alibaba International Digital Commercial Group \\
        \texttt{\{yuzhan.yzj, xiaokejunkejun.xia\}@alibaba-inc.com}
       }

\begin{document}
\maketitle
\renewcommand{\thefootnote}{}
\footnotetext{$^{*}$Equal contribution.}
\footnotetext{$^{\dagger}$Corresponding author.}
\renewcommand{\thefootnote}{\arabic{footnote}}
\begin{abstract}
In e-commerce, LLM agents show promise for shopping tasks such as recommendations, budget management, and bundle deals, where accurately capturing user preferences from long-horizon conversations is critical.
However, progress is limited by two key challenges: (1) the absence of benchmarks for evaluating long-term preference-aware shopping tasks, and (2) the lack of fine-grained supervision for shopping agent training. 
To fill the benchmark gap, we introduce \textsc{Shopping Companion Bench}, a novel benchmark comprising two shopping tasks that require cross-session preference memory, grounded in a product pool of over 1.2 million real-world items.
Our analysis further identifies two major sources of failure on this benchmark: cascading errors caused by preference hallucination, and insufficient verification of product attributes against user requirements.
To address these failure modes, we design annotation-free, tool-wise rewards that provide process supervision for each tool call, alleviating reward sparsity in long-horizon tasks.
Experimental results demonstrate that even state-of-the-art models such as GPT-5 achieve success rates below 70\%, highlighting the difficulty of our benchmark. Notably, our fine-tuned lightweight 4B model consistently outperforms strong baselines in both preference capture and task performance, suggesting the effectiveness of our reward design.
\end{abstract}

\section{Introduction}

\begin{figure*}[t!]
\begin{center}
\includegraphics[width=\linewidth]{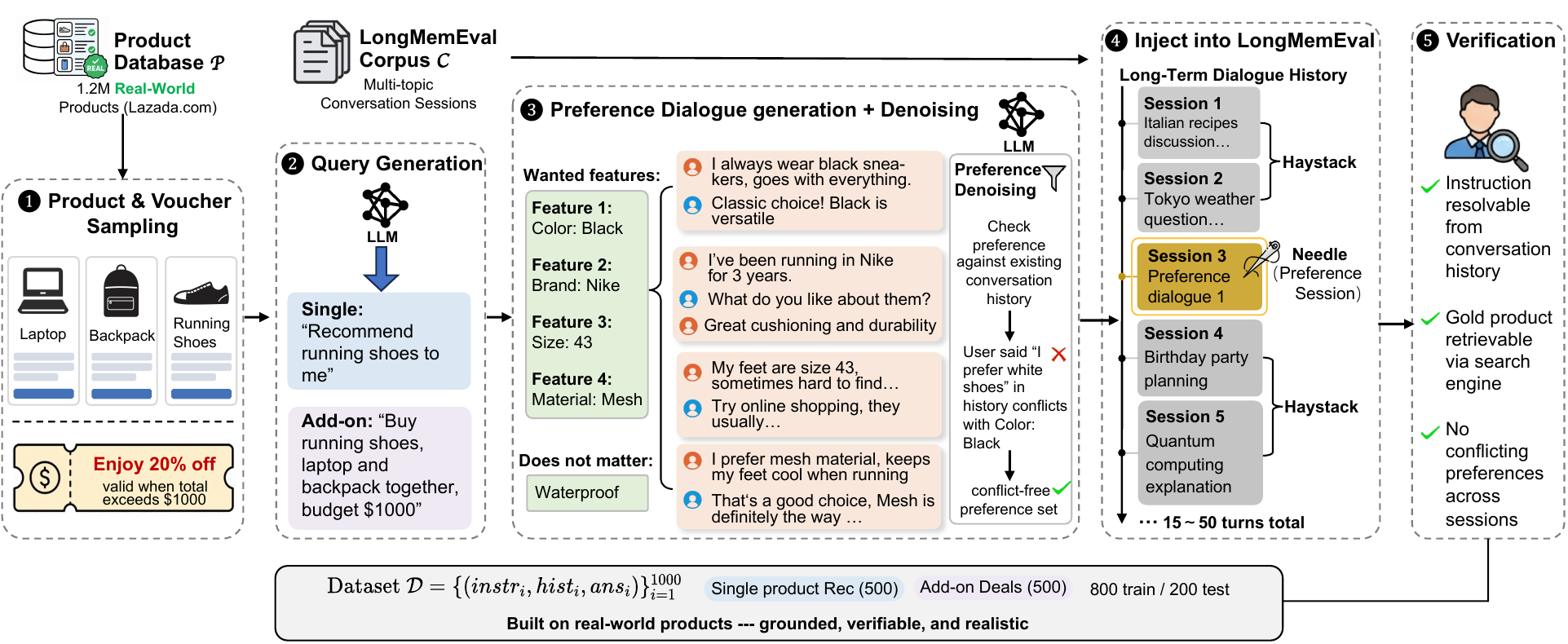}
\caption{Overview of the synthesis pipeline for constructing \textsc{Shopping Companion Bench}. Starting from 1.2M real-world products and the LongMemEval corpus, we sample products and vouchers, generate single-product and add-on-deal instructions, synthesize preference dialogues with denoising, inject them as preference ``needles'' into long-term conversation histories, and verify that each instance is resolvable, searchable, and free of conflicting preferences.}
\label{fig:benchmark_pipeline}
\end{center}
\vspace{-1.5em}
\end{figure*}

\begin{table}[t]
\centering
\footnotesize
\resizebox{\columnwidth}{!}{%
\begin{tabular}{@{}lccc@{}}
\toprule
\textbf{Benchmark} & \textbf{Task} & \textbf{LTM} & \textbf{Intera.} \\
\midrule
WebShop {\scriptsize\citep{yao2022webshop}}            & \textcolor{green!60!black}{\checkmark} & \textcolor{red}{\ding{55}} & \textcolor{red}{\ding{55}} \\
LongMemEval {\scriptsize\citep{wu2024longmemeval}}      & \textcolor{red}{\ding{55}}  & \textcolor{green!60!black}{\checkmark} & \textcolor{red}{\ding{55}} \\
ShoppingBench {\scriptsize\citep{wang2025shoppingbench}} & \textcolor{green!60!black}{\checkmark} & \textcolor{red}{\ding{55}} & \textcolor{red}{\ding{55}} \\
ShopSimulator {\scriptsize\citep{wang2026shopsimulator}} & \textcolor{green!60!black}{\checkmark} & \textcolor{red}{\ding{55}} & \textcolor{green!60!black}{\checkmark} \\
\textbf{Ours}                                           & \textcolor{green!60!black}{\checkmark} & \textcolor{green!60!black}{\checkmark} & \textcolor{green!60!black}{\checkmark} \\
\bottomrule
\end{tabular}%
}
\caption{Comparison of our benchmark with existing works. \textbf{Task} denotes downstream task evaluation. \textbf{LTM} signifies long-term memory evaluation. \textbf{Intera.} indicates whether the evaluation protocol supports multi-turn interaction, user corrections, and preference updates.}
\label{tab:benchmark_comparison}
\vspace{-1.5em}
\end{table}

Large language model (LLM) agents are increasingly used in e-commerce tasks for recommendation, budget management, and bundle deals. Unlike question answering, shopping assistance is inherently action-centric: agents must interact with large product databases, inspect attributes, enforce constraints, and revise decisions based on intermediate observations. A central challenge is capturing long-term user preferences, which are often expressed only implicitly across extended conversations, such as brand aversions, size history, and material preferences.

However, to the best of our knowledge, no prior benchmark integrates cross-session preference memory, real-world shopping tasks, and user preferences correction during evaluation. As shown in Table~\ref{tab:benchmark_comparison}, WebShop~\citep{yao2022webshop} focuses on single-session search without long-term memory. LongMemEval~\citep{wu2024longmemeval} evaluates memory across sessions but lacks downstream tasks. ShoppingBench~\citep{wang2025shoppingbench} covers diverse shopping intents but does not support long-term memory or user correction. ShopSimulator~\citep{wang2026shopsimulator} enables multi-turn interaction, yet relies on static, expert-summarized preferences instead of conversational memory. 

To address this gap, we introduce \textsc{Shopping Companion Bench}, a benchmark constructed from 1.2M real-world products sourced from Lazada.com. It comprises two task types: single-product recommendation and add-on deal recommendation. Preference dialogues are synthesized and injected as ``needles'' into long conversation histories from LongMemEval, followed by human verification to ensure consistency. The final benchmark contains 1,000 instances, evenly split across the two task types.

Beyond benchmark construction, a complementary challenge is how to train shopping agents for long-horizon tool use. Agents must retrieve preferences from memory, search products, verify attributes, and generate recommendations that satisfy both explicit requirements and implicit preferences. Final-outcome supervision alone is often too coarse. Our analysis of a GPT-5-based agent reveals systematic failures: early mistakes arise from missing or hallucinated preferences due to poor memory use, while later errors stem from skipping attribute verification, leading to plausible but invalid recommendations.

Motivated by this, we propose \emph{tool-wise rewards}, a fine-grained supervision signal for intermediate tool use. For each tool call, we assign a continuous reward based on how many returned results match the user’s ground-truth preferences. Derived automatically from tool traces and synthesized task settings, these rewards provide dense feedback without manual annotation and alleviate the sparsity of outcome-only reinforcement learning.

We evaluate both proprietary and open-source LLMs, and further train a lightweight model using our reward scheme. Results show that the benchmark is challenging even for frontier models: GPT-5 is the strongest zero-shot baseline, but its success rate remains below 70\%. More importantly, a 4B model trained with our method substantially outperforms supervised fine-tuning and outcome-only RL, consistently improving both preference grounding and end-to-end success. Further analysis shows that tool-wise rewards encourage more targeted tool use, fewer redundant turns, and shorter responses, improving both effectiveness and efficiency.

In summary, our contributions are threefold:
\begin{itemize}[leftmargin=*, topsep=2pt, itemsep=2pt, parsep=0pt]
\item \textbf{Benchmark.} We introduce \textsc{Shopping Companion Bench}, a novel benchmark for jointly evaluating cross-session preference memory, end-to-end shopping over real-world products, and interactive multi-turn execution with user simulation.
\item \textbf{Tool-wise rewards.} We propose annotation-free tool-wise rewards for memory and product tool calls, providing process-level supervision beyond outcome-only training.
\item \textbf{Experiments and analysis.} Extensive experiments show that the benchmark is challenging even for strong proprietary and open-source models. Our analysis identifies preference hallucination and missing attribute verification as key failure modes, and demonstrates that our reward design mitigates them, enabling a lightweight 4B model to outperform strong baselines.
\end{itemize}

\section{Related Work}
\label{sec:related_work}
\paragraph{Long-Term Memory.} Long-term memory is a key capability for LLM agents in long-horizon interactions. Existing work mainly augments models with external memory through retrieval-based pipelines or explicit read/write memory modules \citep{lewis2020retrieval,karpukhin2020dense,guu2020retrieval,zhong2024memorybank,packer2023memgpt,xu2025mem,chhikara2025mem0,rasmussen2025zep}. However, prior studies show that these systems still struggle with multi-session preference tracking and efficient use of context \citep{wu2024longmemeval}. A main reason is that memory is often treated as a post-hoc component rather than optimized for downstream task success. Agentic Memory \citep{yu2026agentic} moves in this direction by training memory operations as tool actions with step-wise supervision. Our work is aligned with this direction, but focuses on a more challenging application: long-horizon shopping assistance, where preference errors can propagate through multi-stage decision-making.

\paragraph{Shopping Agent.} Shopping agents require grounding in large product catalogs and the ability to satisfy user constraints over multi-turn interactions. Prior work has studied conversational product search and recommendation, focusing on preference elicitation, clarification, and ranking \citep{zhang2018towards,bi2019conversational,zou2022learning}. More recent benchmarks evaluate broader shopping abilities, including realistic search workflows, shopping knowledge, and end-to-end task completion \citep{li2025wizard,wang2025shoppingbench,jin2024shopping,wang2025ecomscriptbench}. Nevertheless, existing settings mainly focus on single-session assistance and outcome-level evaluation, leaving long-term cross-session preference modeling largely unexplored. In contrast, we introduce \textsc{Shopping Companion Bench}, a benchmark for long-horizon preference-aware shopping grounded in a large real-world product pool. We further show that failures often arise from preference hallucination and missing product verification, motivating our annotation-free tool-wise rewards for both memory and product tool calls.

\section{Problem Formulation}
\label{sec:problem_formulation}
We formulate the task as a partially observable Markov decision process defined by the tuple $(\mathcal{S}, \mathcal{A}, \mathcal{T}, \mathcal{O}, \mathcal{R})$, consisting of a state space $\mathcal{S}$, an action space $\mathcal{A}$, a transition function $\mathcal{T}$, an observation space $\mathcal{O}$, and a reward function $\mathcal{R}$.

At each time step $t$, the agent receives an observation derived from the underlying state $s_t \in \mathcal{S}$, composed of the conversation context $C_t$, the long-term memory store $\mathcal{M}_t$ capturing user preferences, and the natural language instruction $\mathcal{I}$:
\begin{equation}
s_t = (C_t,\; \mathcal{M}_t,\; \mathcal{I}).
\end{equation}

Given $s_t$ and LLM parameters $\theta$, the agent performs an action $a_t \in \mathcal{A}$, interacts with the environment, and updates the state:
\begin{equation}
(s_{t+1},\; o_{t+1}) = \mathcal{T}(a_t|s_t; \theta).
\end{equation}
This formulation treats memory-based preference capture and shopping assistance as integral components of the same decision process, rather than separate modules.
At the terminal state $s_T$, the task is successful if the agent's final recommendation: (1) satisfies all needs $n \in \mathcal{N}(\mathcal{I})$; (2) matches all preferences $p \in \mathcal{P}(\mathcal{M})$:

\begin{equation}
C_{\mathcal{I}} = \bigwedge_{n \,\in\, \mathcal{N}(\mathcal{I})} \textsc{Satisfy}(s_T,\, n),
\end{equation}

\begin{equation}
C_{\mathcal{M}} = \bigwedge_{p \,\in\, \mathcal{P}(\mathcal{M})} \textsc{Match}(s_T,\, p),
\end{equation}

\begin{equation}
\textsc{Success}(s_T) =
\begin{cases}
1, & \text{if } C_{\mathcal{I}} \wedge C_{\mathcal{M}}, \\[6pt]
0, & \text{otherwise.}
\end{cases}
\end{equation}

\begin{figure*}[ht!]
\begin{center}
\includegraphics[width=\linewidth]{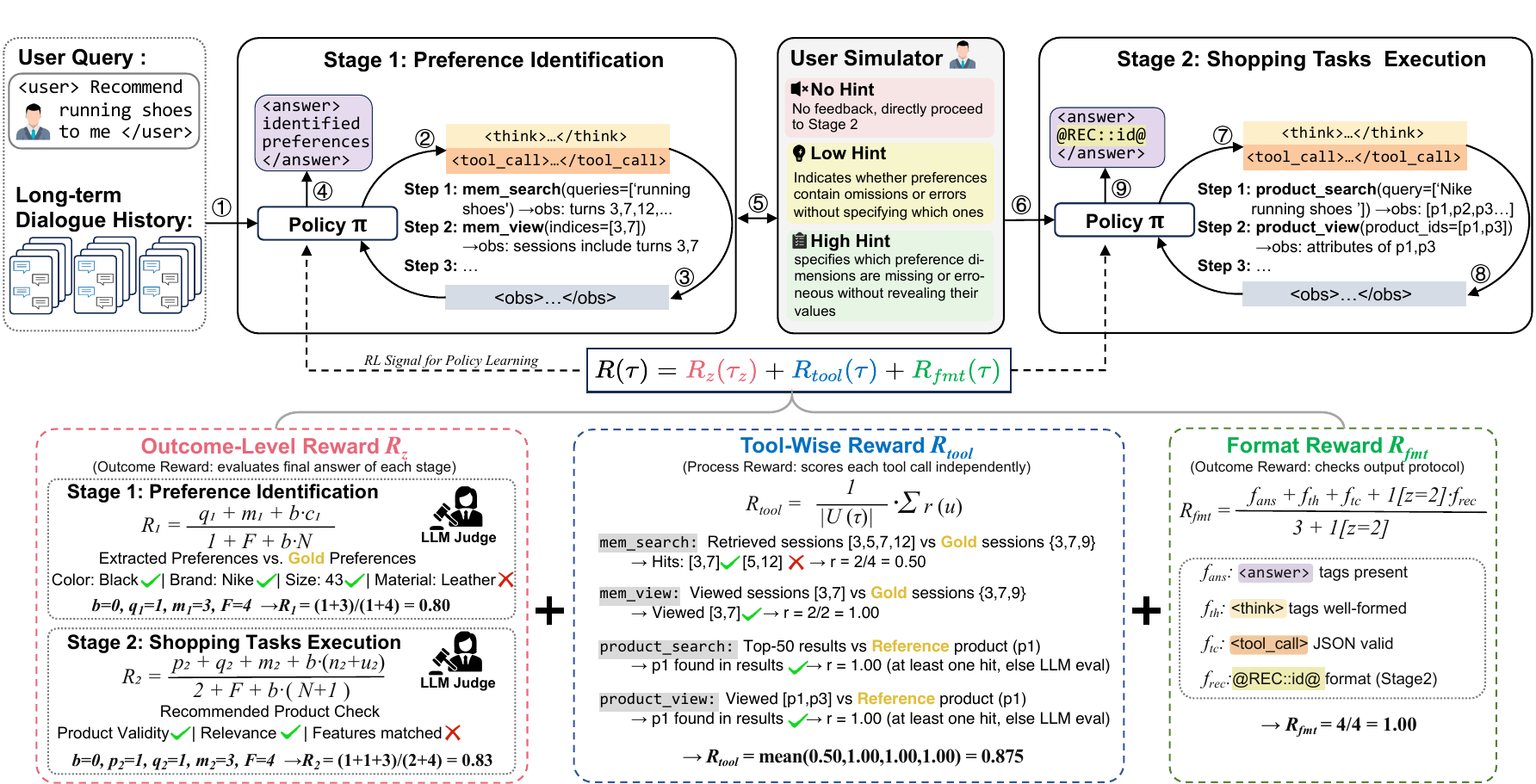}
\caption{Overview of our agent framework, training workflow, and reward design. \textbf{Top}: The agent framework organizes shopping assistance into preference identification and task execution: Stage~1 retrieves and confirms user preferences from long-term memory; during training, the user simulator follows the No-Hint mode and directly passes the identified preferences to Stage~2, while Low-Hint and High-Hint modes are used for evaluation-time correction studies; Stage~2 iteratively searches products and verifies constraints. \textbf{Bottom}: The final reward $R(\tau) = R_z(\tau_z) + R_{\mathrm{tool}}(\tau) + R_{\mathrm{fmt}}(\tau)$, comprising outcome-level reward $R_z$, tool-wise reward $R_{\mathrm{tool}}$ (process-level scoring of each tool invocation for credit assignment), and format reward $R_{\mathrm{fmt}}$ (output protocol compliance).}
\label{fig:rl_reward}
\vspace{-1.5em}
\end{center}

\end{figure*}

\section{Benchmark Construction}
\label{sec:benchmark_construction}

We build \textsc{Shopping Companion Bench} through five steps illustrated in Figure~\ref{fig:benchmark_pipeline}. Inspired by real-world e-commerce scenarios, the benchmark defines two tasks of increasing difficulty:
\begin{itemize}[leftmargin=*, topsep=2pt, itemsep=2pt, parsep=0pt]
\item \textbf{Single-product recommendation} requires retrieving relevant preferences from long-term memory and finding one matching product (\emph{single-hop preference retrieval} + \emph{product search and verification}).
\item \textbf{Add-on deal recommendation} requires retrieving multiple preferences for several products and reasoning over voucher thresholds and budget constraints (\emph{multi-hop preference retrieval} + \emph{multi-product search} + \emph{arithmetic reasoning}).
\end{itemize}

\paragraph{Step 1: Product \& Voucher Sampling.} We source 1,298,797 real-world products from Lazada.com, obtained through a research collaboration with the platform, across broad categories (electronics, fashion, home \& living, etc.), each with searchable fields including title, price, category, brand, description, and structured attributes. For single-product tasks, we sample one target product; for add-on deals, we additionally sample voucher constraints and a bundle of compatible products under a shared budget.

\paragraph{Step 2: Query Generation.} Given the sampled products (and voucher, if applicable), an LLM generates a natural-language shopping instruction. For single-product tasks, the instruction requests a recommendation matching implicit preferences. For add-on deals, it specifies a multi-product purchase under budget constraints.

\paragraph{Step 3: Preference Dialogue Generation \& Denoising.} From the target product attributes, we partition features into \emph{wanted features} and \emph{does-not-matter features}. An LLM then converts the wanted features into multi-turn preference dialogues that naturally embed these preferences in conversation.
To ensure consistency with the target user's existing conversation history, we perform a \emph{preference denoising} step: given the wanted and does-not-matter feature lists, an LLM iterates over each session of the randomly selected user in the LongMemEval corpus and checks whether any existing statement conflicts with the generated preferences. Conflicting preferences are either removed or regenerated until a conflict-free preference set is obtained.

\paragraph{Step 4: Injection into LongMemEval.} To better reflect real-world complexity, we inject these verified preference dialogues as a ``needle'' session into users’ multi-topic conversation histories from LongMemEval~\citep{wu2024longmemeval} under license MIT, thereby creating a needle-in-a-haystack setting~\citep{grover1997quantum}. The surrounding sessions (e.g., discussions about Italian recipes, weather-related questions, and birthday planning) serve as distractors in the haystack. The resulting long-term dialogue history spans 15--50 turns, with preference-relevant evidence being sparse and embedded within unrelated conversations.

\paragraph{Step 5: Verification.} Each instance is double-checked by two e-commerce experts to ensure that: (1) the instruction must be resolvable from the conversation history; (2) the reference product must be retrievable using the product search engine; and (3) no conflicting preferences may remain across sessions. This process ensures that evaluation failures reflect limitations of the agent rather than ambiguity in the dataset. The benchmark includes 1,000 instructions, evenly divided between single-product recommendation and add-on deals, with 800 training and 200 test examples. More statistics are provided in Appendix~\ref{appendix:benchmark_statistics}.

\paragraph{Evaluation Metrics.} Because valid recommendation need not exactly match the reference products, alternative products may be correct if they satisfy all constraints and preferences, we adopt the LLM-as-Judge paradigm~\citep{zheng2023judgingllmasajudgemtbenchchatbot} to evaluate semantic correctness beyond exact matching:

\begin{itemize}[leftmargin=*, topsep=0pt, itemsep=0pt, parsep=0pt]
\item \textbf{Accuracy (Acc.)}: fraction of reference preference attributes correctly retrieved by the agent.
\item \textbf{Success Rate (Succ.)}: fraction of test cases satisfying all criteria: (1) correct product count, (2) user need satisfaction, (3) consistency with long-term preferences, and (4) budget feasibility (for add-on deals).
\end{itemize}

To verify robustness, we evaluate using five diverse LLM judges and observe a high OPC ($0.9547 \pm 0.0289$), indicating low sensitivity to the choice of judge model. Human meta-evaluation on 200 samples by 8 annotators further confirms reliability, showing strong human agreement (iPAR = 0.8950) and high alignment between the LLM judge and human consensus (PAR = 0.9497). Details shown in Appendix~\ref{sec:eval_robustness}.

\begin{figure*}[ht!]
\centering
\begin{minipage}[t]{0.60\textwidth}
    \centering
    \includegraphics[width=\textwidth]{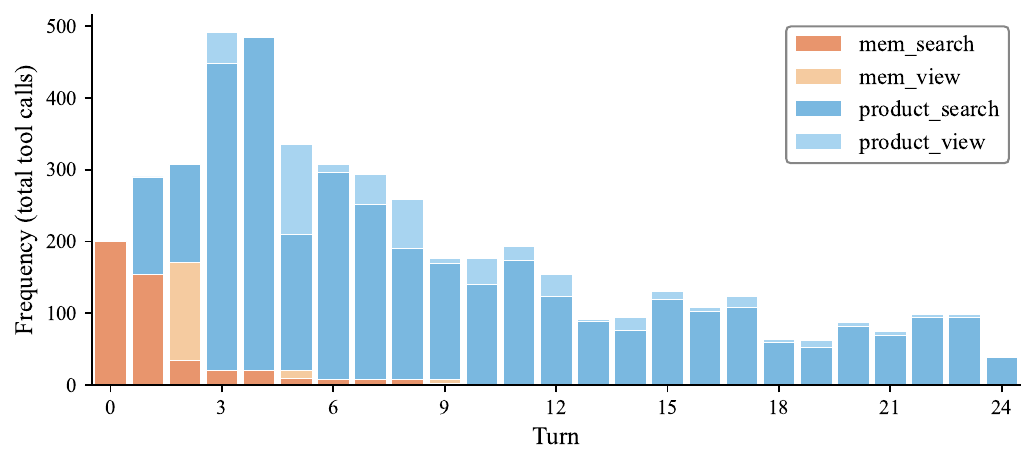}
\end{minipage}\hspace{0.01\textwidth}%
\begin{minipage}[t]{0.38\textwidth}
    \centering
    \includegraphics[width=\textwidth]{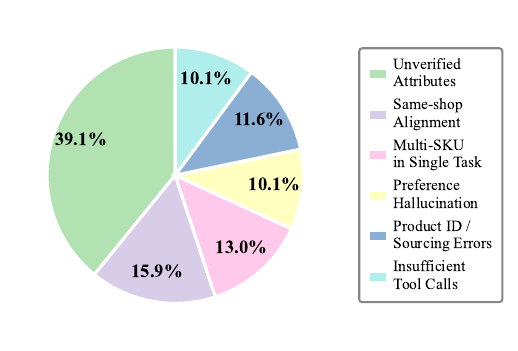}
\end{minipage}
\vspace{-1.0em}
\caption{Preliminary analysis of the GPT-5 zero-shot agent. \textbf{Left}: Tool-call frequency by turn over 200 trajectories. The agent exhibits a two-phase behavior: an initial memory-retrieval stage (Turns 0--2), followed by an iterative search--view stage (Turns 3+). \textbf{Right}: Distribution of failure modes among 69 failed trajectories. The two largest categories are \emph{Unverified Attributes} and \emph{Same-Shop Alignment}.}
\label{fig:behavior_analysis}
\vspace{-1.5em}
\end{figure*}

\section{Training Method}
\label{sec:shopping_companion}
With the benchmark in place, we perform a preliminary analysis of a GPT-5 agent to identify the main causes of failure in long-horizon shopping tasks. By analyzing trajectory-level behaviors and attributing failures, we uncover two key issues: cascading errors caused by preference hallucination and insufficient verification of product attributes against user requirements. Based on these findings, we develop an annotation-free tool-wise reward design for targeted process-level supervision. 

This section first introduces the agent framework (Sec.~\ref{sec:shopping_two_stage}), then presents the preliminary analysis (Sec.~\ref{sec:preliminary_analysis}), and finally details the reward design (Sec.~\ref{sec:shopping_reward_design}).


\subsection{Agent Framework}
\label{sec:shopping_two_stage}


Our agent is formulated as a tool-using policy $\pi$ with access to four tools---memory search, memory view, product search, and product view---and follows a two-stage workflow, as illustrated in the top panel of Figure~\ref{fig:rl_reward}. Detailed descriptions of the four tools are provided in Appendix~\ref{appendix:tool_interface}.

\noindent \textbf{Stage 1: Preference Identification.}
Given a user query and long-term dialogue history, the agent uses memory tools to retrieve relevant sessions, recognize implicit shopping preferences and ask for confirmation.

\noindent \textbf{Stage 2: Shopping Tasks Execution.}
Given the user query,  recognized preferences in stage 1, and the user correction (depends on the user simulation), the agent searches for products, inspects product attributes, and verifies both explicit user requirements and implicit preferences before producing a special structured recommendation.

\noindent \textbf{User Simulator.}
Between Stage 1 and Stage 2, a user simulator may provide corrective feedback in one of three modes:
\begin{itemize}[nosep,leftmargin=1.2em]
\item \textbf{No-Hint}: No feedback is provided, and the agent proceeds directly to Stage~2.
\item \textbf{Low-Hint}: The simulator indicates whether the extracted preferences contain omissions or errors, without specifying which ones.
\item \textbf{High-Hint}: The simulator specifies which preference keys are missing or erroneous, without revealing the ground-truth values.
\end{itemize}

\subsection{Preliminary Analysis}
\label{sec:preliminary_analysis}
We randomly sampled 200 instances (100 per task type), ran zero-shot rollouts with GPT-5, and analyzed the 69 failed trajectories evaluated by \texttt{Succ.} metrics.

\paragraph{Behavior patterns.}
Across the trajectories, GPT-5 exhibits a consistent two-phase strategy, as shown in the left panel of Figure~\ref{fig:behavior_analysis}: it first performs brief memory retrieval in the initial few turns, and then quickly shifts into a prolonged product search--view loop. This pattern suggests that preference identification is typically compressed into an early and fragile stage. As a result, once the agent forms an incorrect preference hypothesis, subsequent product searches are systematically steered in the wrong direction, producing cascading errors that are difficult to recover from later. Notably, the two-stage agent framework is not the cause of the poor performance. In Sec.~\ref{sec:user_simulator_exp}, we find that the agent performs far worse than the oracle under both the single-stage and two-stage settings, with the former performing even worse than the latter.

\paragraph{Failure attribution.}
Our attribution analysis of the failed trajectories further shows that the dominant source of failure is not merely retrieval errors, but insufficient verification during execution. The two largest categories are \emph{Unverified Attributes} and \emph{Same-Shop Alignment}. In these cases, the agent recommends products without verifying their key attributes or seller information against user requirements. Together with \emph{Preference Hallucination} and the closely related issue of insufficient memory inspection, these results indicate that the main weaknesses lie in (i) early preference errors that trigger trajectory-level cascades and (ii) inadequate verification of attributes and seller information before the final recommendation. Detailed statistics, a taxonomy, and representative case studies are provided in Appendix~\ref{appendix:behavior_patterns} and Appendix~\ref{appendix:error_taxonomy}.




\subsection{Reward Function Design}
\label{sec:shopping_reward_design}

Therefore, we introduce tool-wise rewards to not only reinforce correct tool-use behaviors, but also provide earlier supervision along the trajectory, rather than relying solely on terminal feedback.

As shown in the bottom panel of Figure~\ref{fig:rl_reward}, the RL signal consists of three complementary components: (1)~an \emph{outcome-level reward} $R_z$ that evaluates the final answer quality of each stage, (2)~a \emph{tool-wise process reward} $R_{\mathrm{tool}}$ that scores each intermediate tool call for finer-grained credit assignment, and (3)~a \emph{format reward} $R_{\mathrm{fmt}}$ that enforces output protocol compliance. All terms are differentiated by task type ($b=0$ for single-product, $b=1$ for add-on-deals).

\paragraph{Outcome-Level Reward $R_z$.}
Let $z\in\{1,2\}$ denote the current stage, $F$ the number of required preference attributes, and (for add-on-deals, $b{=}1$) $N$ the reference bundle size. Reward prompts are in Appendix~\ref{app:reward_prompts}.

\noindent\emph{Stage-1 (Preference Identification):}
The judge returns query relevance $q_1\in\{0,1\}$, matched preference-attribute count $m_1\in[0,F]$, and (when $b{=}1$) matched product-count $c_1\in[0,N]$:
\begin{equation}
\label{eq:stage1_reward_main}
R_1(\tau_1)
=
\frac{q_1 + m_1 + b\,c_1}{1 + F + b\,N}.
\end{equation}

\noindent\emph{Stage-2 (Shopping Tasks Execution):}
The judge checks product validity $p_2\in\{0,1\}$, relevance $q_2\in\{0,1\}$, and matched preference-attribute count $m_2\in[0,F]$.
For add-on-deals ($b{=}1$), it additionally returns matched product-count $n_2\in[0,N]$ and budget feasibility $u_2\in\{0,1\}$:
\begin{equation}
\label{eq:stage2_reward_main}
R_2(\tau_2)
=
\frac{p_2 + q_2 + m_2 + b\,(n_2+u_2)}{2 + F + b\,(N+1)}.
\end{equation}
\paragraph{Tool-Wise Process Reward $R_{\mathrm{tool}}$.}
Let $\mathcal{U}(\tau)$ be the set of tool calls in trajectory $\tau$, and $r(u)\in[0,1]$ the per-call fine-grained reward score:

For \textbf{memory tools}, let $\mathcal{T}_{\mathrm{mem}}$ contain memory search/view calls, $\mathcal{S}_{\text{gold}}$ denote the set of gold preference sessions, and $I_u$ be the session indices retrieved or viewed by tool call $u$:
\begin{equation}
r(u)=\frac{|\{i \in I_u : \text{sess}(i) \in \mathcal{S}_{\text{gold}}\}|}{|I_u|},
\quad u\in\mathcal{T}_{\mathrm{mem}}.
\end{equation}

For \textbf{product tools}, let $\mathcal{T}_{\mathrm{prod}}$ contain product search/view calls, $\mathcal{P}_{\text{ref}}$ denote the reference product IDs, and $P_u$ be the products retrieved or viewed by tool call $u$:
\begin{equation}
r(u)=\frac{|\{p \in P_u : p \in \mathcal{P}_{\text{ref}}\}|}{|P_u|},
\quad u\in\mathcal{T}_{\mathrm{prod}}.
\end{equation}
When exact ID matching is unavailable, we fall back to LLM-based semantic matching against reference features.

The overall tool-wise reward aggregates over all tool calls via their mean:
\begin{equation}
\label{eq:toolwise_mean}
R_{\mathrm{tool}}(\tau)
=
\begin{cases}
\frac{1}{|\mathcal{U}(\tau)|}\sum\limits_{u\in\mathcal{U}(\tau)} r(u),
& |\mathcal{U}(\tau)|>0,\\[6pt]
0, & \text{otherwise}.
\end{cases}
\end{equation}

\begin{table*}[t]
\centering
\small
\setlength{\tabcolsep}{4pt}
\renewcommand{\arraystretch}{1.0}

\begin{tabular}{llcccccc}
\toprule
\multirow{2}{*}{\textbf{Category}} & \multirow{2}{*}{\textbf{Model}} &
\multicolumn{2}{c}{\textbf{Single Product}} &
\multicolumn{2}{c}{\textbf{Add-on Deals}} &
\multicolumn{2}{c}{\textbf{Average}} \\
\cmidrule(lr){3-4}\cmidrule(lr){5-6}\cmidrule(lr){7-8}
& & \textbf{Acc.(\%)} & \textbf{Succ.(\%)} & \textbf{Acc.(\%)} & \textbf{Succ.(\%)} & \textbf{Acc.(\%)} & \textbf{Succ.(\%)} \\
\midrule
\multirow{4}{*}{Closed} 
& GPT-5 & 82.0 & 75.0 & \textbf{66.0} & \textbf{54.0} & \textbf{74.0} & \textbf{64.5} \\
& GPT-4.1 & 88.0 & 78.0 & 39.0 & 24.0 & 63.5 & 51.0 \\
& GPT-4o & 79.0 & 72.0 & 41.0 & 26.0 & 60.0 & 49.0 \\
& Qwen3-Max & 80.0 & 72.0 & 35.0 & 24.0 & 57.5 & 48.0 \\

\midrule
\multirow{6}{*}{Open} 
& DeepSeek-R1 & 64.0 & 57.0 & 40.0 & 24.0 & 52.0 & 40.5 \\
& Kimi-K2-Instruct & 81.0 & 75.0 & 23.0 & 14.0 & 52.0 & 44.5 \\
& Gemma-3-27B & 56.0 & 50.0 & 18.0 & 11.0 & 37.0 & 30.5 \\
& Qwen3-Next-80B-A3B & 63.0 & 57.0 & 29.0 & 18.0 & 46.0 & 37.5 \\
& Qwen3-30B-A3B & 60.0 & 53.0 & 21.0 & 13.0 & 40.5 & 33.0 \\
& Qwen3-4B & 49.0 & 44.0 & 11.0 & 6.0 & 30.0 & 25.0 \\

\midrule
\multirow{5}{*}{\centering Ours} & Qwen3-4B-LoRA & 82.0 & 72.0 & 42.0 & 31.0 & 62.0 & 51.5  \\[0.5ex]
& \begin{tabular}[c]{@{}l@{}}Qwen3-4B-LoRA + RL \\ (Outcome-level)\end{tabular} & \underline{89.0} & \underline{81.0} & 50.0 & 38.0 & 69.5 & 59.5  \\[0.5ex]
& \begin{tabular}[c]{@{}l@{}}Qwen3-4B-LoRA + RL \\ (Outcome-level\&Tool-wise Reward)\end{tabular} & \textbf{90.0} & \textbf{84.0} & \underline{55.0} & \underline{43.0} & \underline{72.5} & \underline{63.5}  \\
\bottomrule
\end{tabular}
\caption{Main results on \textsc{Shopping Companion Bench}. Acc.\ measures preference grounding in Stage~1; Succ. measures final recommendation success in Stage~2. The \textbf{best} and \underline{second-best} results are marked.}
\label{tab:main_results}
\vspace{-1.5em}
\end{table*}

\paragraph{Format Reward $R_{\mathrm{fmt}}$.}
A lightweight format reward stabilizes structured generation.
Four binary indicators check protocol compliance: $f_{\mathrm{ans}}$ (answer tags present), $f_{\mathrm{th}}$ (thinking tags well-formed), $f_{\mathrm{tc}}$ (tool-call JSON parsable), and $f_{\mathrm{rec}}$ (recommendation conforms to required schema, Stage~2 only):
\begin{equation}
\label{eq:fmt_reward_main}
R_{\mathrm{fmt}}(\tau)
=
\frac{
f_{\mathrm{ans}} + f_{\mathrm{th}} + f_{\mathrm{tc}} + \mathbb{I}[z{=}2]\cdot f_{\mathrm{rec}}
}{
3+\mathbb{I}[z{=}2]
}.
\end{equation}

\paragraph{Final Reward.}
The total reward for a training sample at stage $z$ sums all three components with unit weights:
\begin{equation}
\label{eq:final_reward}
R(\tau)
=
R_{z}(\tau_z)
+
R_{\mathrm{tool}}(\tau)
+
R_{\mathrm{fmt}}(\tau).
\end{equation}

\section{Experiments}
\subsection{Experimental Setup}

\paragraph{Implementation Details.}
We reject sample 2,948 step-level examples with GPT-4.1 for SFT, and then further train the SFT model with RL on 800 instances from the training set. Further implementation details are provided in Appendix~\ref{appendix:Experimental Implementation}.

\paragraph{Baselines.}
We compare closed-source LLMs (GPT~\citep{OpenAI2024MemoryControls} and Qwen3-Max~\citep{qwen3_max_official}) and open-source LLMs (Qwen3~\citep{yang2025qwen3}, DeepSeek-R1~\cite{guo2025deepseek}, Kimi-K2-Instruct~\cite{team2025kimi}, and Gemma-3~\citep{gemma3_2025}) under zero-shot settings. For our fine-tuned model, we report progressive improvements from (1) Qwen3-4B + LoRA fine-tuning, to (2) Qwen3-4B + LoRA + outcome-level reward RL, and (3) Qwen3-4B + LoRA + outcome-level reward + tool-wise reward. During training, the user-simulator interface follows the No-Hint mode; Low-Hint and High-Hint are only used in the evaluation-time user intervention study.

\subsection{Main Results}

Table~\ref{tab:main_results} reports results on both single-product recommendation and add-on deals.

\paragraph{Closed-source LLMs.}
Closed-source models achieve strong performance on single-product tasks. However, performance drops substantially on add-on deals, with success rates between 24.0\% and 54.0\%, indicating that multi-product coordination and constraint satisfaction remain challenging even for large-scale models.
\paragraph{Open-source LLMs.}
We evaluate open-source models across diverse families (Qwen, DeepSeek, Moonshot, Google). Among them, Kimi-K2-Instruct achieves 75.0\% single-product Succ.---comparable to GPT-5---but drops sharply to 14.0\% on add-on deals, revealing that single-product capability does not transfer to multi-product coordination. DeepSeek-R1 shows more balanced performance (57.0\%/24.0\%) but still substantially trails GPT-5. Smaller Qwen3 models (4B--30B) and Gemma-3-27B perform poorly overall, particularly on add-on deals ($\leq$13.0\% Succ.).

\paragraph{Ours.}
Our trained agent yields consistent improvements. LoRA fine-tuning substantially boosts the 4B backbone from 25.0\% to 51.5\% average Succ., and outcome-level reward RL further improves to 59.5\%. Adding tool-wise rewards reaches 90.0\% Acc. and 84.0\% Succ. on single products and 55.0\% Acc. and 43.0\% Succ. on add-on deals (72.5\% / 63.5\% average). This surpasses all open-source baselines---including models 7$\times$ larger---and approaches the best closed-source model (GPT-5), validating the effectiveness of end-to-end optimization with fine-grained credit assignment over memory and product tool calls.

\subsection{Ablation Studies}
\label{sec:tool_wise_analysis}

\paragraph{User Simulator.} 
\label{sec:user_simulator_exp}
We conduct an evaluation-time ablation on the user simulator with LLM (Appendix~\ref{appendix:user_simulator}). \textit{Oracle} provides the ground-truth preference-relevant sessions as context and serves as an upper bound. \textit{One-Stage} performs preference identification and shopping assistance end-to-end. \textit{Two-Stage} separates them, with three feedback modes after Stage~1: \textit{No Hint} (no correction), \textit{Low Hint} (only whether errors/omissions exist), and \textit{High Hint} (which preference dimensions are wrong or missing, but not their values).

As shown in Table~\ref{tab:user_simulator}, \textit{One-Stage} achieves only 52.5\% average success (32.0\% on Add-on Deals), indicating that jointly handling preference grounding and shopping overwhelms the model. \textit{Two-Stage (No Hint)} improves to 64.5\%, and user feedback further helps: \textit{Low Hint} and \textit{High Hint} reach 68.5\% and 70.0\%, respectively. The remaining gap to \textit{Oracle} indicates that preference grounding remains a key bottleneck.


\paragraph{Tool-wise Reward.}
To analyze the effect of tool-wise supervision, we compare outcome-level and outcome-level \& tool-wise reward training strategies from both trajectory-level and behavioral perspectives.

\begin{table}[htbp]
\centering
\small
\setlength{\tabcolsep}{8pt}
\renewcommand{\arraystretch}{1.2}
\begin{tabular}{l r r r}
\hline
    \textbf{Strategy}          & \textbf{Single} & \textbf{Add-on} & \textbf{Avg.} \\
\hline
Oracle            & \textbf{85.0} & \textbf{73.0} & \textbf{79.0} \\
One-Stage         & 73.0 & 32.0 & 52.5 \\
Two-Stage (No Hint)  & 75.0 & 54.0 & 64.5 \\
Two-Stage (Low Hint)   & 78.0 & 59.0 & 68.5 \\
Two-Stage (High Hint)  & \underline{80.0} & \underline{60.0} & \underline{70.0} \\
\hline
\end{tabular}
\caption{Success rates (\%) for workflow variants and user simulator. The backbone model is GPT-5; No-Hint corresponds to proceeding without corrective feedback.}
\label{tab:user_simulator}
\vspace{-1.5em}
\end{table}

\begin{figure}[htbp]
    \centering
    \includegraphics[width=\linewidth]{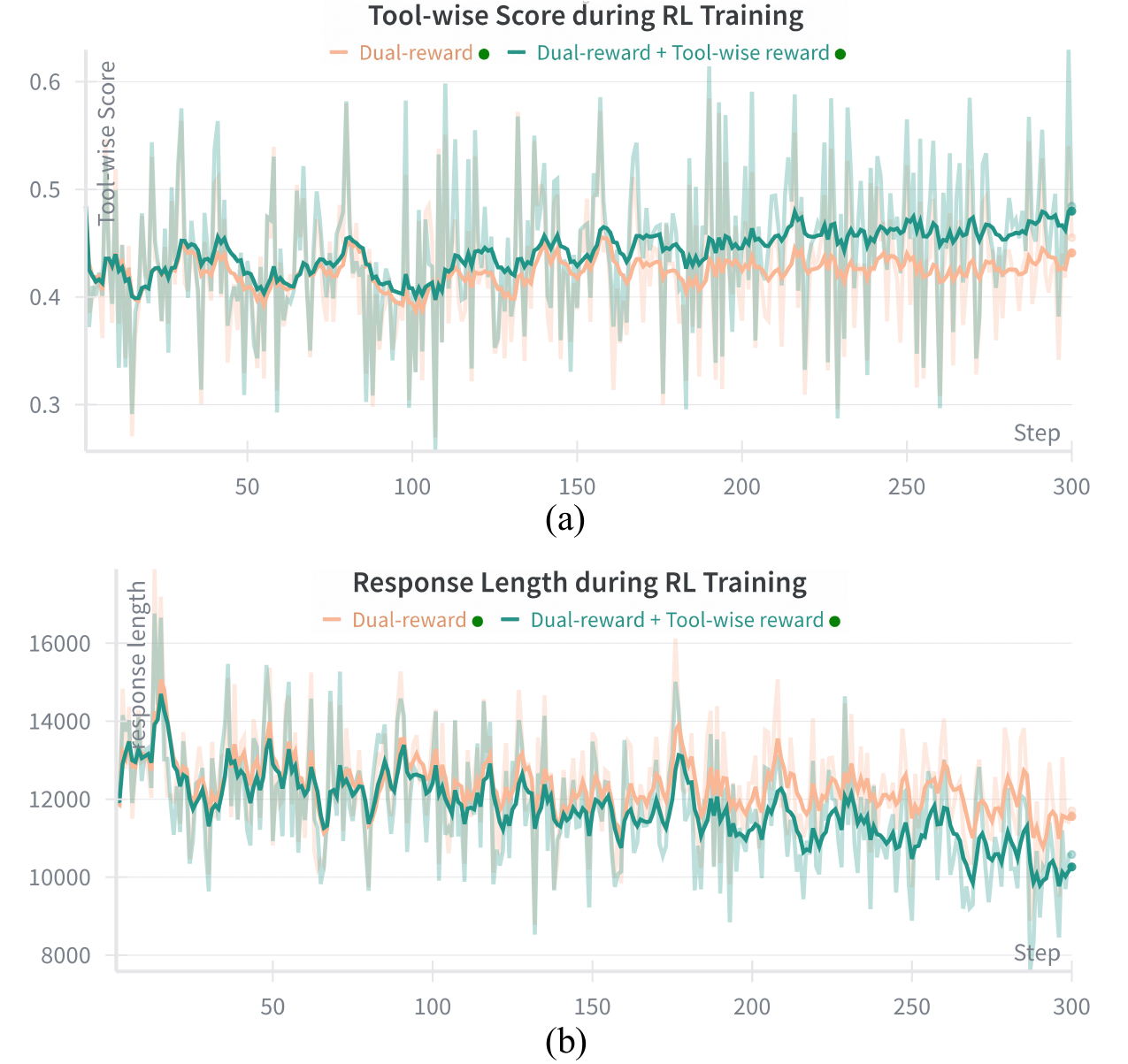}
    \caption{Tool-wise reward effects: (a) tool-wise score and (b) response length over training steps (solid=smoothed, translucent=raw).}
    \label{fig:toolwise_ablation}
\vspace{-1.5em}
\end{figure}

\textbf{Tool utilization quality.}
As shown in Figure~\ref{fig:toolwise_ablation}(a), incorporating tool-wise reward consistently increases the averaged tool-wise score throughout training, with a widening gap in later stages. This indicates improved credit assignment for intermediate tool decisions, encouraging more relevant memory and product retrieval behaviors.

\textbf{Efficiency and verbosity control.}
Figure~\ref{fig:toolwise_ablation}(b) shows that the tool-wise variant produces shorter responses and exhibits a clearer downward trend over training. This suggests that step-level supervision not only improves tool correctness but also reduces unnecessary long-form generations, leading to more efficient trajectories.

In addition, Table~\ref{tab:reward_analysis} reveals that the outcome-level \& tool-wise variant demonstrates fewer redundant turns, more targeted tool usage, and shorter responses, further validating that granular reward signals effectively shape agent behavior beyond terminal success optimization.

\begin{table}[htbp]
\centering
\small
\setlength{\tabcolsep}{2.5pt}
\renewcommand{\arraystretch}{1.0}
\begin{tabular}{lccc}
\toprule
\textbf{Strategy} & \textbf{Turns} & \textbf{Tool Uses} & \textbf{Resp. Len.} \\
\midrule
Outcome-level &9.82 &9.17 &10485.39 \\
Outcome-level + Tool-wise &8.89 &8.47 &10068.83 \\
\bottomrule
\end{tabular}
\caption{Behavioral metrics comparison (mean values).}
\label{tab:reward_analysis}
\vspace{-1.5em}
\end{table}

\section{Conclusion}


We introduce \textsc{Shopping Companion Bench}, a product-grounded benchmark for evaluating shopping agents that must recover long-term user preferences from cross-session conversations and apply them to real-world recommendation tasks. Our results show that this setting remains challenging even for strong LLMs: GPT-5 achieves 64.5\% zero-shot success, with most failures arising from insufficient verification and other tool-calling errors. Guided by this analysis, we propose annotation-free tool-wise rewards that provide fine-grained supervision over memory and product tool calls, complementing outcome-level feedback. Training a lightweight 4B agent with this objective improves both preference grounding and task success, reaching 63.5\% average success and approaching GPT-5 performance. 

\newpage
\section*{Limitations}
While our work presents a product-grounded benchmark and an agent training method for preference-grounded shopping assistance, we acknowledge the following limitations:

\paragraph{Difficulty of budget-constrained multi-item tasks.} Although our trained agent consistently outperforms open-source baselines across both tasks, performance on the add-on deals task remains relatively low for all evaluated models. This task requires the agent to jointly reason over budget constraints, inter-product compatibility, and diverse user preferences—a combinatorial challenge that proves substantially harder than single-item recommendation. The results suggest that current approaches, including ours, still have significant room for improvement in handling constrained, multi-product optimization within conversational settings.

\paragraph{Generalizability of tool-wise reward design.} Our reinforcement learning strategy with outcome-level and tool-wise rewards is designed around the specific tool set used in our shopping tasks. Extending this approach to other domains or broader tool-augmented agent settings may require non-trivial adaptation, as decomposing sparse task-level feedback into fine-grained, per-tool supervision depends on domain-specific considerations. How to systematically and scalably design tool-wise rewards in more general settings remains an open question.

\section*{Ethical Considerations}
This work studies e-commerce assistance with long-term preference memory, raising privacy and fairness concerns.
First, cross-session memory may expose sensitive information if mishandled; deployments should minimize stored content, enforce retention policies, encrypt data, and provide user controls to inspect, correct, and delete memories.
Second, preference inference can enable unwanted profiling or amplify biases; systems should avoid inferring sensitive attributes without explicit user intent and should be audited for disparate outcomes.
Finally, recommendations can shape spending behavior; agents should communicate uncertainty, avoid manipulative framing, and prioritize user-aligned constraints (e.g., budget and safety).
Regarding artifact use, all existing resources employed in this work---including LongMemEval (MIT license), Pyserini, and pretrained language models---are used strictly for academic research consistent with their stated intended use. Product data was obtained through an authorized research collaboration with the Lazada platform and is used solely for academic research purposes. Our benchmark and associated code are likewise intended solely for research purposes.

\bibliography{ref}

\newpage

\appendix
\section{Supplemental Details For Our Benchmark}
\label{appendix:benchmark}
\subsection{Basic Statistics}
\label{appendix:benchmark_statistics}
We provide detailed visualizations of the benchmark data distribution.

\begin{figure*}[h]
\centering
\includegraphics[width=0.85\linewidth]{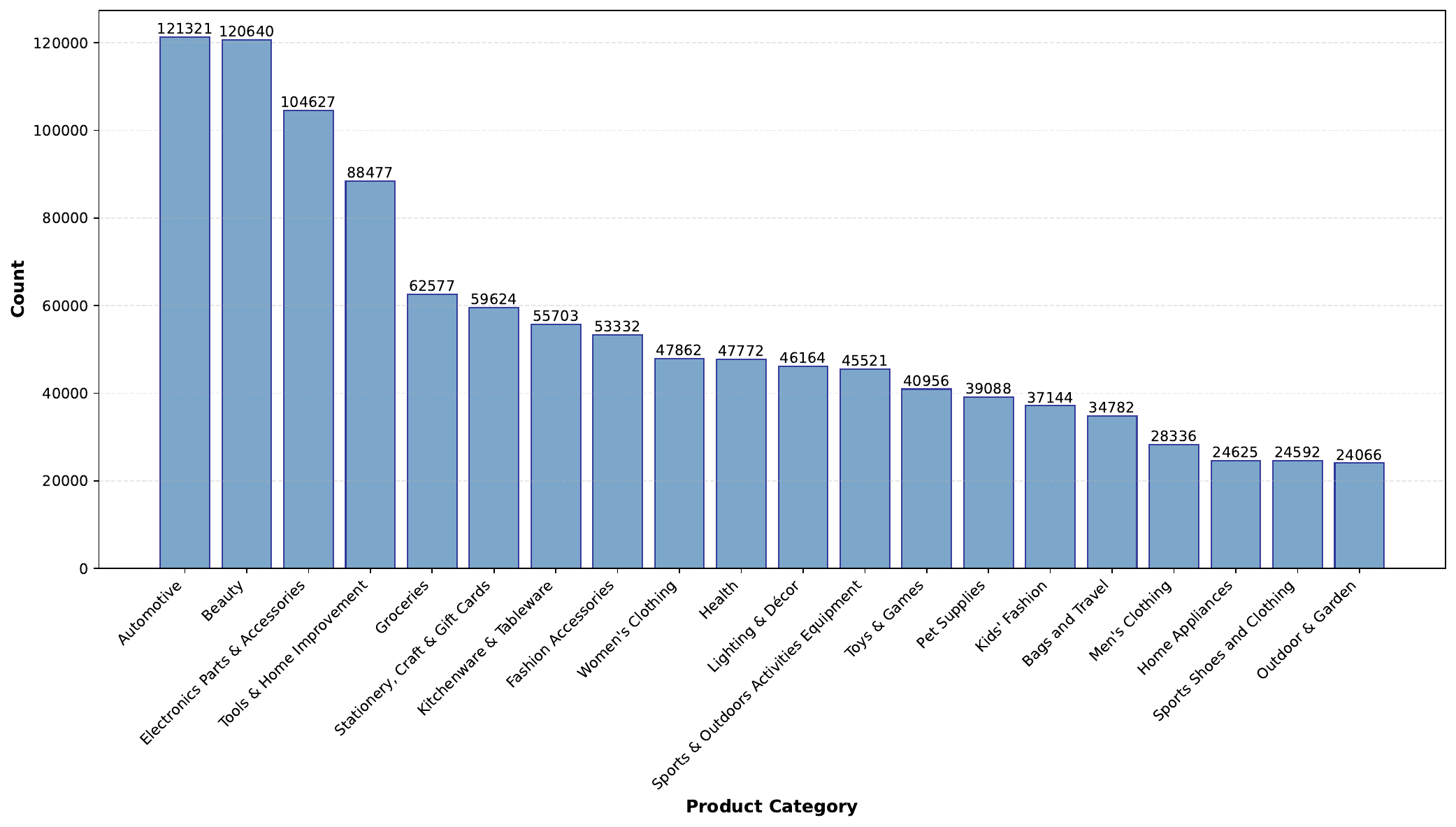}
\vspace{-0.5em}
\caption{Distribution of Product Category (Top 20)}
\label{fig:category_distribution}
\vspace{0.3em}

\includegraphics[width=0.85\linewidth]{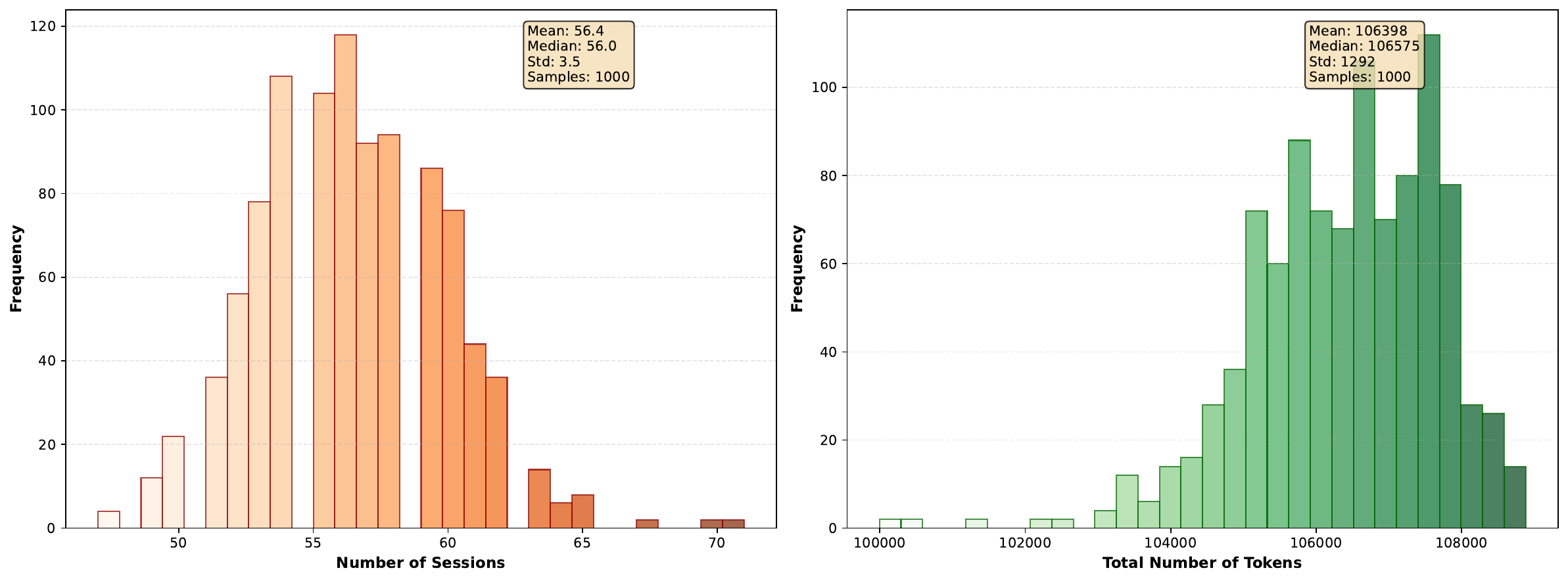}
\vspace{-0.5em}
\caption{Distribution of Sessions and Total Tokens}
\label{fig:haystack_distributions}
\vspace{0.3em}

\includegraphics[width=0.85\linewidth]{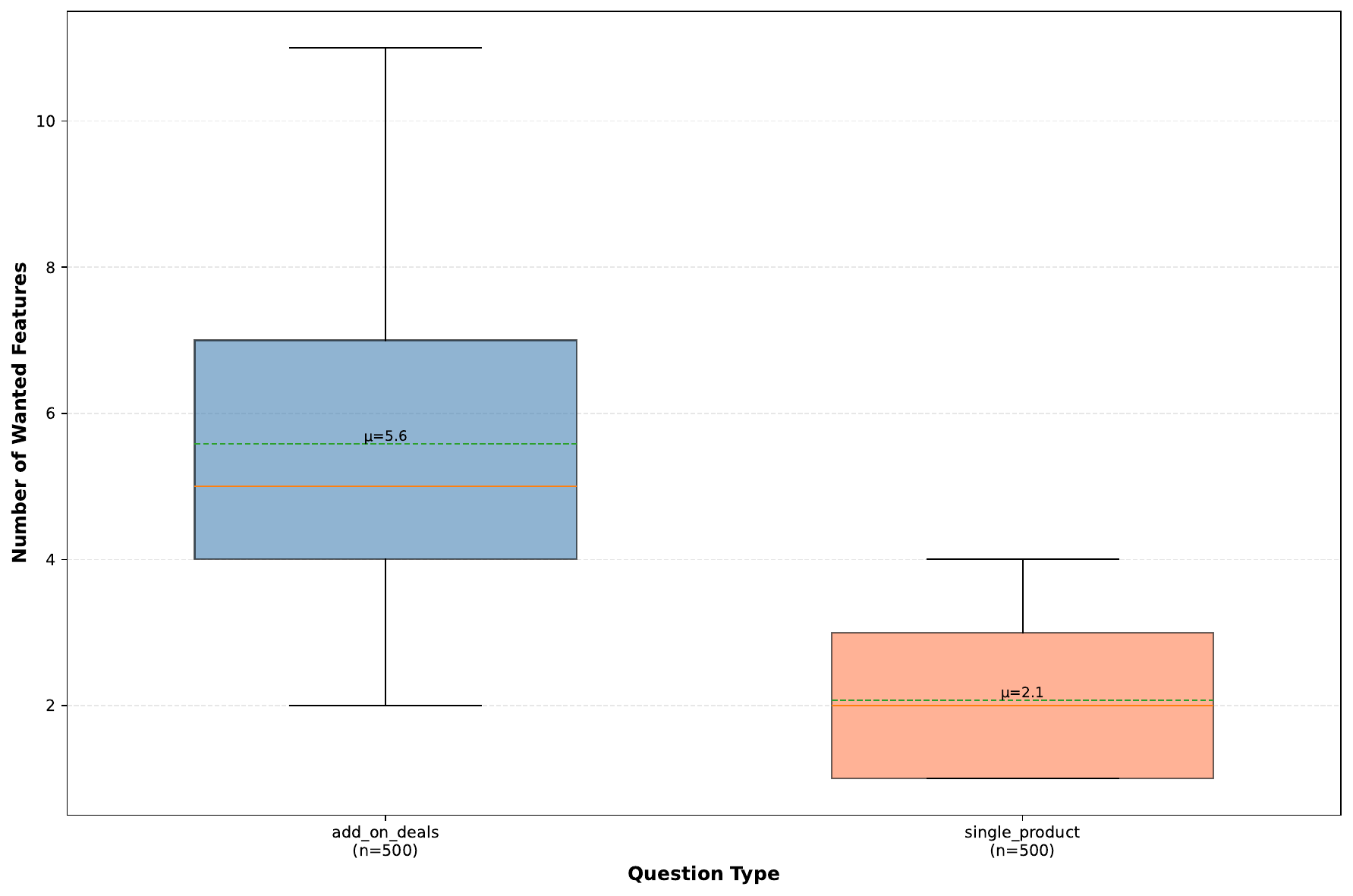}
\vspace{-0.5em}
\caption{Distribution of Wanted Features}
\label{fig:wanted_features_distribution}
\end{figure*}

Figure~\ref{fig:category_distribution} presents the product category distribution for the top 20 categories, revealing broad but imbalanced coverage across product domains. The largest categories include \textit{Automotive} (121,321), \textit{Beauty} (120,640), and \textit{Electronics Parts \& Accessories} (104,627), followed by categories such as \textit{Tools \& Home Improvement} and \textit{Grocery}. Even lower-ranked categories among the top 20 still contain tens of thousands of instances, indicating substantial category diversity while preserving realistic long-tail characteristics.

Figure~\ref{fig:haystack_distributions} shows the distributions of session count and total token count per user. The number of sessions is concentrated around the mid-50s, with a mean of 56.4, a median of 56.0, and a standard deviation of 3.3, suggesting relatively consistent user interaction depth across samples. The total token count per user is centered around 106k tokens (mean: 106,380; median: 106,575; std: 1,311), indicating that the benchmark contains long interaction histories and therefore poses a nontrivial long-context understanding challenge.

Figure~\ref{fig:wanted_features_distribution} illustrates the number of desired preference attributes for different task types. The \textit{add-on deals} setting involves substantially richer preference constraints, with an average of 5.6 attributes and an interquartile range of roughly 4 to 7, whereas \textit{single-product} tasks are much simpler, with an average of 2.1 attributes and most cases falling between 1 and 3. This contrast reflects the varying complexity of user preference specification across task types and highlights the benchmark's ability to evaluate recommendation under both simple and compositional preference conditions.

\subsection{Long-Term Memory Construction}
\label{appendix:long_term_memory_construction}
The preferences-evidence dialogue session generation prompt is shown in Figure~\ref{fig:dialogue_generation_prompt}. The user instruction generation prompts for Single Product and Add-on Deals are shown in Figure~\ref{fig:single_generation} and Figure~\ref{fig:add_on_deals_generation}, respectively.

In the actual production process, to mitigate the bias introduced by any single model, we evenly distributed the synthesis tasks among GPT-5, GPT-4.1, Qwen3-Max, and Kimi-K2-Instruct when synthesizing user instructions and dialogue sessions. Consequently, each LLM was responsible for generating 250 instructions and 250 sessions.

\begin{figure*}[h]
\centering
\begin{tcolorbox}[
    enhanced,
    colback=blue!3!white,
    colframe=blue!40!black,
    fonttitle=\bfseries\sffamily,
    title={\small Dialogue Generation Prompt},
    coltitle=white,
    attach boxed title to top left={yshift=-2mm, xshift=3mm},
    boxed title style={colback=blue!40!black, sharp corners},
    sharp corners,
    boxrule=0.5pt,
    left=3pt,
    right=3pt,
    top=8pt,
    bottom=3pt,
    arc=0pt
]
\begin{lstlisting}[
    language=Python,
    basicstyle=\footnotesize\ttfamily\color{red!60!black},
    keywordstyle=\color{red!60!black},
    stringstyle=\color{red!60!black},
    commentstyle=\color{red!60!black}\itshape,
    numberstyle=\tiny\color{red!60!black},
    showstringspaces=false,
    breaklines=true,
    columns=flexible,
    keepspaces=true,
    tabsize=4,
    xleftmargin=0pt,
    xrightmargin=0pt,
    frame=none
]
<instructions>
You are a professional scriptwriter.

**Input:**
- <product_name>: The product being discussed.
- <features>: Features of the product.

**Tasks:**
1. Choose {number_of_features} meaningful and easy-to-understand features from <features> as **wanted features** and treat the rest as **does not matter features**.
2. Craft a realistic user-assistant dialogue.

**Dialogue Rules:**
1. **Role Clarity**:
   - The user proposes a life scenario where the product would be needed.
   - The assistant knows well the scenario and the product.
2. **User Behavior**:
   - Describe the usage scenario and explain why they want the product.
   - Should not voluntarily propose any features, but passively respond to the those that the assistant proposes.
   - Clearly states whether a feature is **wanted** or **does not matter** through personal experiences.
   - Confirm all the wanted features before the conversation ends, but do not buy the product now.
3. **Assistant Strategy**:
   - Do not make the conversation too salesy. The assistant's goal is to help the user analyze the demand and make a purchase decision.
   - Ask about one aspect at a time to avoid overwhelming the user.
   - Proactively ask clarification questions while the user clarifies the needs.
   - Discuss only features that appear in <features>; exclude features mentioned only in the <product_name>.
4. **Conversation Flow**:
   - Natural, coherent, diverse, and contextually consistent.
   - Reorganize and sequence the attributes logically to reflect a realistic discovery process.
   - Only output utterances, no non-verbal actions.
</instructions>

<product_name>
{product_name}
</product_name>

<features>
{features}
</features>

<output_format>
Return as a valid JSON object with "wanted_features", "does_not_matter_features", and "dialogue" keys:
- "wanted_features": a valid JSON array of strings, each string is a feature that the user wants.
- "does_not_matter_features": a valid JSON array of strings, each string is a feature that the user does not matter.
- "dialogue": a valid JSON array of JSON objects, each object has "role" and "content" keys.
Example format:
```json
{{
    "wanted_features": ["f1: v1", "f2: v2", ...],
    "does_not_matter_features": ["f3: v3", "f4: v4", ...],
    "dialogue": [
        {{"role": "user", "content": "..."}},
        {{"role": "assistant", "content": "..."}},
        ...
    ]
}}
```
</output_format>
"""
\end{lstlisting}
\end{tcolorbox}
\caption{Dialogue Generation Prompt}
\label{fig:dialogue_generation_prompt}
\end{figure*}

\begin{figure*}[h]
\centering
\begin{tcolorbox}[
    enhanced,
    colback=blue!3!white,
    colframe=blue!40!black,
    fonttitle=\bfseries\sffamily,
    title={\small User Instruction Generation Prompt of Single Product Rec},
    coltitle=white,
    attach boxed title to top left={yshift=-2mm, xshift=3mm},
    boxed title style={colback=blue!40!black, sharp corners},
    sharp corners,
    boxrule=0.5pt,
    left=3pt,
    right=3pt,
    top=8pt,
    bottom=3pt,
    arc=0pt
]
\begin{lstlisting}[
    language=Python,
    basicstyle=\footnotesize\ttfamily\color{red!60!black},
    keywordstyle=\color{red!60!black},
    stringstyle=\color{red!60!black},
    commentstyle=\color{red!60!black}\itshape,
    numberstyle=\tiny\color{red!60!black},
    showstringspaces=false,
    breaklines=true,
    columns=flexible,
    keepspaces=true,
    tabsize=4,
    xleftmargin=0pt,
    xrightmargin=0pt,
    frame=none
]
# Task
Write a query to search for a product on the e-commerce platform.

# Product Information
{product_name}

# Important Notes
- Only use the basic product category name (no more than 5 tokens) from the title.
- Don't mention specific product attributes.
- Don't repeat the product title in the query.

Generate the query without any other text:
\end{lstlisting}
\end{tcolorbox}
\caption{User Instruction Generation Prompt of Single Product Rec}
\label{fig:single_generation}
\end{figure*}

\begin{figure*}[h]
\centering
\begin{tcolorbox}[
    enhanced,
    colback=blue!3!white,
    colframe=blue!40!black,
    fonttitle=\bfseries\sffamily,
    title={\small User Instruction Generation Prompt of Add-On Deals},
    coltitle=white,
    attach boxed title to top left={yshift=-2mm, xshift=3mm},
    boxed title style={colback=blue!40!black, sharp corners},
    sharp corners,
    boxrule=0.5pt,
    left=3pt,
    right=3pt,
    top=8pt,
    bottom=3pt,
    arc=0pt
]
\begin{lstlisting}[
    language=Python,
    basicstyle=\footnotesize\ttfamily\color{red!60!black},
    keywordstyle=\color{red!60!black},
    stringstyle=\color{red!60!black},
    commentstyle=\color{red!60!black}\itshape,
    numberstyle=\tiny\color{red!60!black},
    showstringspaces=false,
    breaklines=true,
    columns=flexible,
    keepspaces=true,
    tabsize=4,
    xleftmargin=0pt,
    xrightmargin=0pt,
    frame=none
]
# Task
Write a query to search for a product bundle (include multiple products) on the e-commerce platform.

# Product Information
{product_names}

# Deal Information
Voucher Details: {voucher}
Budget: ${budget}

# Important Notes
- The query MUST include the voucher details and budget.
- Clearly state the multiple products in the bundle and use "," or "and" to connect them.
- The query should be like this: "Product bundle includes: X, Y, and Z. Voucher ... Budget ...".
- Only use the basic product category name (no more than 5 tokens) from the title.
- Don't mention specific product attributes.
- Don't repeat the product title in the query.

Generate the query without any other text:
\end{lstlisting}
\end{tcolorbox}
\caption{User Instruction Generation Prompt of Add-On Deals}
\label{fig:add_on_deals_generation}
\end{figure*}

\subsection{Evaluation Methods Building}
\label{appendix:evaluate_methods_building}
To accurately evaluate the diverse responses of LLMs, we employ an expert-written prompt to instruct LLM as the correctness judge. We present the full prompts in Figure~\ref{fig:single_evaluation} and Figure~\ref{fig:add_on_deals_evaluation}. Since our benchmark spans two shopping tasks over a large-scale product catalog, each task type involves distinct evaluation considerations; we therefore design separate prompts for each task to enable the model to handle detailed edge cases as expert evaluators would. Specifically, for single-product recommendation, the evaluator checks whether the recommended product is relevant to the user query and contains all wanted features, matching semantically rather than requiring exact wording. For bundle-deal scenarios involving multiple products, the evaluator additionally verifies that the recommended set covers all products specified in the user query, with each product satisfying its corresponding wanted features.

\begin{figure*}[h]
\centering
\begin{tcolorbox}[
    enhanced,
    colback=blue!3!white,
    colframe=blue!40!black,
    fonttitle=\bfseries\sffamily,
    title={\small Single Product Rec Evaluation Prompt},
    coltitle=white,
    attach boxed title to top left={yshift=-2mm, xshift=3mm},
    boxed title style={colback=blue!40!black, sharp corners},
    sharp corners,
    boxrule=0.5pt,
    left=3pt,
    right=3pt,
    top=8pt,
    bottom=3pt,
    arc=0pt
]
\begin{lstlisting}[
    language=Python,
    basicstyle=\footnotesize\ttfamily\color{red!60!black},
    keywordstyle=\color{red!60!black},
    stringstyle=\color{red!60!black},
    commentstyle=\color{red!60!black}\itshape,
    numberstyle=\tiny\color{red!60!black},
    showstringspaces=false,
    breaklines=true,
    columns=flexible,
    keepspaces=true,
    tabsize=4,
    xleftmargin=0pt,
    xrightmargin=0pt,
    frame=none
]
You are a professional query-product matching evaluator.

**Task:**
Given a user query, wanted features, and a recommended product, answer "yes" ONLY if BOTH conditions are met:
1. The recommended product is relevant to the user query
2. The recommended product contains ALL wanted features

**Matching Rules:**
- Compare features comprehensively: consider name, attributes, and options of the recommended product
- Match features semantically (exact wording not required, meaning must align)
- Ignore attribute order and extra attributes in the recommended product
- Answer "no" if ANY wanted feature is missing or unclear

**Default to "no" when uncertain.**

** User Query:**
{user_query}

** User Wanted Features:**
{wanted_features}

** Recommended Product:**
{recommended_product}

Output "yes" or "no" without any other text.
\end{lstlisting}
\end{tcolorbox}
\caption{Single Product Rec Evaluation Prompt}
\label{fig:single_evaluation}

\end{figure*}

\begin{figure*}[h]
\centering
\begin{tcolorbox}[
    enhanced,
    colback=blue!3!white,
    colframe=blue!40!black,
    fonttitle=\bfseries\sffamily,
    title={\small Add-On Deals Evaluation Prompt},
    coltitle=white,
    attach boxed title to top left={yshift=-2mm, xshift=3mm},
    boxed title style={colback=blue!40!black, sharp corners},
    sharp corners,
    boxrule=0.5pt,
    left=3pt,
    right=3pt,
    top=8pt,
    bottom=3pt,
    arc=0pt
]
\begin{lstlisting}[
    language=Python,
    basicstyle=\footnotesize\ttfamily\color{red!60!black},
    keywordstyle=\color{red!60!black},
    stringstyle=\color{red!60!black},
    commentstyle=\color{red!60!black}\itshape,
    numberstyle=\tiny\color{red!60!black},
    showstringspaces=false,
    breaklines=true,
    columns=flexible,
    keepspaces=true,
    tabsize=4,
    xleftmargin=0pt,
    xrightmargin=0pt,
    frame=none
]
You are a professional query-product matching evaluator.

** Task:**
Given:
- user query: include multiple products
- wanted features: include features that correspond to each product in the user query
- recommended products
Answer "yes" ONLY if BOTH conditions are met:
1. The recommended products include ALL the products in the user query
2. For each product in the user query, the recommended product contains ALL the corresponding wanted features

** Matching rules:**
- Compare features comprehensively: consider name, attributes, and options of the recommended product
- Match features semantically (exact wording not required, meaning must align)
- Ignore the product order in the user query and the recommended products
- Ignore the attribute order and extra attributes in the recommended product
- Answer "no" if ANY wanted feature is missing or unclear

**Default to "no" when uncertain.**

** User Query:**
{user_query}

** User Wanted Features:**
{wanted_features}

** Recommended Products:**
{recommended_products}

Output "yes" or "no" without any other text.
\end{lstlisting}
\end{tcolorbox}
\caption{Add-On Deals Evaluation Prompt}
\label{fig:add_on_deals_evaluation}
\end{figure*}

\section{Supplemental Details For Agent Framework}
\subsection{Tool Interface}
\label{appendix:tool_interface}

The agent interacts with two external resources: a long-term conversation memory and a product catalog. We expose these resources through four tools, separating coarse retrieval from fine-grained inspection in both memory and product spaces.

\paragraph{\texttt{mem\_search}.}
This tool performs query-based retrieval over the user's long-term conversation history. Given one or more natural-language search queries, it returns a ranked list of potentially relevant dialogue sessions or snippets. The tool is intended for broad recall: the agent uses it to locate candidate sessions that may contain preference evidence, such as brand dislikes, size history, material constraints, or previous shopping experiences. Because the returned snippets may be incomplete or only partially relevant, \texttt{mem\_search} should not be treated as sufficient evidence for final preference extraction.

\paragraph{\texttt{mem\_view}.}
This tool provides detailed access to full dialogue sessions identified by \texttt{mem\_search}. Given session indices, it returns the corresponding conversation content, allowing the agent to verify whether an inferred preference is explicitly supported by the user's past utterances. The intended use is confirmation rather than discovery: after \texttt{mem\_search} retrieves candidate sessions, \texttt{mem\_view} helps the agent distinguish genuine preferences from misleading snippets, avoid preference hallucination, and extract preference attributes with supporting evidence.

\paragraph{\texttt{product\_search}.}
This tool retrieves candidate products from the product catalog. Given a product query, and optionally constraints such as shop ID or price range, it returns a list of products with basic metadata, including product IDs, titles, prices, categories, and seller information. The tool is designed for candidate generation during shopping execution. For add-on-deal tasks, the optional shop constraint is especially important because all recommended products may need to satisfy same-shop voucher requirements. However, the search results typically contain limited product attributes, so they should be followed by product inspection before recommendation.

\paragraph{\texttt{product\_view}.}
This tool inspects one or more candidate products returned by \texttt{product\_search}. Given product IDs, it returns detailed product information, including descriptions and structured attributes when available. The agent uses \texttt{product\_view} to verify whether candidate products satisfy explicit instruction constraints and implicit memory-derived preferences. This verification step is critical because product titles alone often under-specify attributes such as material, compatibility, safety properties, or bundled components. Skipping \texttt{product\_view} is a major source of unverified-attribute errors in our analysis.

\paragraph{Implementation Details.}
The memory tools (\texttt{mem\_search} and \texttt{mem\_view}) operate over a dense retrieval index built with \texttt{all-MiniLM-L6-v2}~\citep{huggingface-minilm-l6-v2}, a lightweight sentence-transformer that encodes each dialogue turn into a 384-dimensional vector. At indexing time, we segment each user's long-term conversation history into individual turns and encode them offline. At query time, \texttt{mem\_search} encodes the agent's natural-language queries with the same model and retrieves the top-$k$ turns by cosine similarity. \texttt{mem\_view} then fetches the full dialogue session containing the selected turns.

The product tools (\texttt{product\_search} and \texttt{product\_view}) are backed by Pyserini~\citep{lin2021pyserinieasytousepythontoolkit}, an information retrieval toolkit built on top of Apache Lucene. We index the full product catalog (1.29M items) using BM25~\citep{bm25} over concatenated product titles, categories, brands, and descriptions. \texttt{product\_search} issues a BM25 query with optional shop-ID and price-range filters, returning up to 50 candidate products with basic metadata. \texttt{product\_view} retrieves the stored structured attributes and option variants for the requested product IDs directly from the index.

\subsection{Two-Stage Agent}
\label{appendix:two_stage_agent_prompts}
The system prompts for the two-stage agentic framework are shown in Figure~\ref{fig:Stage 1 System Prompt} and Figure~\ref{fig:Stage 2 System Prompt}.

\noindent \textbf{Stage 1 -- Preference Identification:} Given a product search query, this agent retrieves relevant memories from the user's dialogue history using memory search and memory view tools to identify purchase preferences. It reasons step-by-step with multi-turn tool calls and outputs the identified preferences through the user-simulator interface.

\noindent \textbf{Stage 2 -- Shopping Task Execution:} Given the search query and the identified preferences from Stage~1, or corrected preferences in evaluation-time correction studies, this agent searches for products that match those preferences. It uses product search and product view tools to verify product attributes, then produces an expert-level Markdown report explaining alignment with user preferences and provides a best-matching recommendation in a special format (\texttt{@REC::product\_id@}).

\subsection{User Simulator}
\label{appendix:user_simulator}
We conduct an evaluation-time ablation comparing five strategies: \textbf{Oracle}, which directly provides ground-truth preference-relevant dialogue sessions as context; \textbf{One-Stage}, which prompts the LLM to perform preference identification and shopping assistance end-to-end without explicit stage separation; and three \textbf{Two-Stage} variants that use distinct prompts for preference identification and shopping assistance. In the two-stage variants, the first stage outputs retrieved memories and identified preferences to the user-simulator interface. \textbf{No Hint} provides no corrective feedback and directly proceeds to Stage~2; this is the mode used during RL training. \textbf{Low Hint} indicates whether the identified preferences contain omissions or errors without specifying which ones, and \textbf{High Hint} specifies which preference dimensions are missing or erroneous without revealing their values. Low-Hint and High-Hint are used only for evaluation-time correction studies. The user simulator prompts are shown in Figure~\ref{fig:user_low_hint} and Figure~\ref{fig:user_high_hint}.

\begin{figure*}[h]
\centering
\begin{tcolorbox}[
    enhanced,
    colback=blue!3!white,
    colframe=blue!40!black,
    fonttitle=\bfseries\sffamily,
    title={\small Stage 1 System Prompt},
    coltitle=white,
    attach boxed title to top left={yshift=-2mm, xshift=3mm},
    boxed title style={colback=blue!40!black, sharp corners},
    sharp corners,
    boxrule=0.5pt,
    left=3pt,
    right=3pt,
    top=8pt,
    bottom=3pt,
    arc=0pt
]
\begin{lstlisting}[
    language=Python,
    basicstyle=\footnotesize\ttfamily\color{red!60!black},
    keywordstyle=\color{red!60!black},
    stringstyle=\color{red!60!black},
    commentstyle=\color{red!60!black}\itshape,
    numberstyle=\tiny\color{red!60!black},
    showstringspaces=false,
    breaklines=true,
    columns=flexible,
    keepspaces=true,
    tabsize=4,
    xleftmargin=0pt,
    xrightmargin=0pt,
    frame=none
]
# Instructions

## Task
Given a product search query, retrieve the user's relevant memories (dialogue history) and identify their purchase preferences from them.

You can complete the task by:
- Retrieve the most relevant memories (dialogue turns) using the "mem_search" tool.
- View the whole dialogue session using the "mem_view" tool.
- Summarize dialogue sessions by date range using the "mem_summarize_by_date" tool.

## Rules
In each turn you can either:
- Think and make one or more tool calls.
- Provide your final answer and terminate the conversation.
You cannot do both at the same time.

You MUST think step by step and make multi-turn tool calls before providing your final answer.

# Output Format

## For thinking and making tool calls
Format the output as follows:
- Reasoning process MUST be within <think></think> tags.
- Tool calls MUST be within <tool_call></tool_call> tags. Each line MUST be a valid JSON object with "name" and "arguments" fields.
- Strictly follow the template below (DO NOT FORGET THE </tool_call> TAG):
```plaintext
<think>...your reasoning process...

</think>
<tool_call>
{{"name": "mem_search", "arguments": {{"queries": ["...", "..."]}}}}
...

</tool_call>
```

## For providing your final answer
Write concise content within <answer></answer> tags as follows:
- List the relevant parts of memories and identified preferences in human-readable format.
- Ask if the above information is sufficient and accurate.
- Strictly follow the template below:
```plaintext
<answer>...your concise content...

</answer>
```
{available_tools}
"""
\end{lstlisting}
\end{tcolorbox}
\caption{Stage 1 System Prompt}
\label{fig:Stage 1 System Prompt}
\end{figure*}

\begin{figure*}[h]
\centering
\begin{tcolorbox}[
    enhanced,
    colback=blue!3!white,
    colframe=blue!40!black,
    fonttitle=\bfseries\sffamily,
    title={\small Stage 2 System Prompt},
    coltitle=white,
    attach boxed title to top left={yshift=-2mm, xshift=3mm},
    boxed title style={colback=blue!40!black, sharp corners},
    sharp corners,
    boxrule=0.5pt,
    left=3pt,
    right=3pt,
    top=8pt,
    bottom=3pt,
    arc=0pt
]
\begin{lstlisting}[
    language=Python,
    basicstyle=\footnotesize\ttfamily\color{red!60!black},
    keywordstyle=\color{red!60!black},
    stringstyle=\color{red!60!black},
    commentstyle=\color{red!60!black}\itshape,
    numberstyle=\tiny\color{red!60!black},
    showstringspaces=false,
    breaklines=true,
    columns=flexible,
    keepspaces=true,
    tabsize=4,
    xleftmargin=0pt,
    xrightmargin=0pt,
    frame=none
]
# Instructions

## Task
Given the product search query and the user's purchase preferences, find products or product bundles that exactly match them.

You can complete the task by:
- Use the "product_search" tool to search for products. Do not recommend any products from your own knowledge base.
- Use the "product_view" tool to view the attributes and options of the products, and then check if they match the user's preferences.
- Obtain up-to-date or domain-specific knowledge from the Internet using the "web_search" tool, and then visit and summarize the webpages using the "web_visit" tool.

## Rules
In each turn you can either:
- Think and make one or more tool calls.
- Provide your final answer and terminate the conversation.
You cannot do both at the same time.

You MUST think step by step and make multi-turn tool calls before providing your final answer.

# Output Format

## For thinking and making tool calls
Format the output as follows:
- Reasoning process MUST be within <think></think> tags.
- Tool calls MUST be within <tool_call></tool_call> tags. Each line MUST be a valid JSON object with "name" and "arguments" fields.
- Strictly follow the template below (DO NOT FORGET THE </tool_call> TAG):
```plaintext
<think>...your reasoning process...

</think>
<tool_call>
{{"name": "product_search", "arguments": {{"query": "...", "shop_id": "...", "price": "..."}}}}
...

</tool_call>
```

## For providing your final answer
Write a expert-level report within <answer></answer> tags as follows:
- Describe how the products or product bundles you found align with user's preferences.
- Enclose the best-matching recommendation in the **special format**: @REC::product_id@ for a single product, or @REC::product_id1,product_id2,...@ for a product bundle.
- The product_id in the **special format** MUST come from the "product_search" tool response.
- May discuss multiple products or bundles, but only use the **special format** once.
- Is formatted with clear, well-structured Markdown.
- Strictly follow the template below:
```plaintext
<answer>...your expert-level report...

</answer>
```
{available_tools}
\end{lstlisting}
\end{tcolorbox}
\caption{Stage 2 System Prompt}
\label{fig:Stage 2 System Prompt}
\end{figure*}

\begin{figure*}[h]
\centering
\begin{tcolorbox}[
    enhanced,
    colback=blue!3!white,
    colframe=blue!40!black,
    fonttitle=\bfseries\sffamily,
    title={\small User Simulator Prompt for Low Hint},
    coltitle=white,
    attach boxed title to top left={yshift=-2mm, xshift=3mm},
    boxed title style={colback=blue!40!black, sharp corners},
    sharp corners,
    boxrule=0.5pt,
    left=3pt,
    right=3pt,
    top=8pt,
    bottom=3pt,
    arc=0pt
]
\begin{lstlisting}[
    language=Python,
    basicstyle=\footnotesize\ttfamily\color{red!60!black},
    keywordstyle=\color{red!60!black},
    stringstyle=\color{red!60!black},
    commentstyle=\color{red!60!black}\itshape,
    numberstyle=\tiny\color{red!60!black},
    showstringspaces=false,
    breaklines=true,
    columns=flexible,
    keepspaces=true,
    tabsize=4,
    xleftmargin=0pt,
    xrightmargin=0pt,
    frame=none
]
# Task
Compare a hypothesis answer (generated by an AI assistant) against a reference products, and output one or more labels indicating issues.

# Inputs
- reference: one or more products. Each product includes a name and a set of required features.
- hypothesis: the assistant's answer to be evaluated.

# Definitions
- missing: The hypothesis fails to fully cover the reference.
  - Missing any entire product, or
  - Missing any required feature of any product, or
  - Only giving generic statements that do not confirm required features.
- wrong: The hypothesis introduces information not present in the reference.
  - Adds external products not in the reference, or
  - Claims features that are not listed in the reference, or
  - States information that contradicts the reference.
- all matched: The hypothesis exactly matches the reference.
  - Every product and its required features are present.
  - Every required feature is correctly matched to the reference.

# Rules
- Treat matching semantically: synonyms and paraphrases count if they clearly refer to the same product/feature.
- Every product and its required features must be present.
- Ignore details not specified in the reference (e.g., price, seller, shipping, warranty) unless explicitly included as required features.
- If both missing and wrong apply, output both labels.
- Do not include explanations, reasoning, or any extra text.

# Reference
{reference}

# Hypothesis
{hypothesis}

# Output
Return exactly one of: "missing", "wrong", or "all matched".
\end{lstlisting}
\end{tcolorbox}
\caption{User Simulator Prompt for Low Hint}
\label{fig:user_low_hint}
\end{figure*}

\begin{figure*}[h]
\centering
\begin{tcolorbox}[
    enhanced,
    colback=blue!3!white,
    colframe=blue!40!black,
    fonttitle=\bfseries\sffamily,
    title={\small User Simulator Prompt for High Hint},
    coltitle=white,
    attach boxed title to top left={yshift=-2mm, xshift=3mm},
    boxed title style={colback=blue!40!black, sharp corners},
    sharp corners,
    boxrule=0.5pt,
    left=3pt,
    right=3pt,
    top=8pt,
    bottom=3pt,
    arc=0pt
]
\begin{lstlisting}[
    language=Python,
    basicstyle=\footnotesize\ttfamily\color{red!60!black},
    keywordstyle=\color{red!60!black},
    stringstyle=\color{red!60!black},
    commentstyle=\color{red!60!black}\itshape,
    numberstyle=\tiny\color{red!60!black},
    showstringspaces=false,
    breaklines=true,
    columns=flexible,
    keepspaces=true,
    tabsize=4,
    xleftmargin=0pt,
    xrightmargin=0pt,
    frame=none
]
# Task
Compare a hypothesis answer (generated by an AI assistant) against a reference products, and output the miss or wrong feature names.

# Inputs
- reference: one or more products. Each product includes a name and a set of required features.
- hypothesis: the assistant's answer to be evaluated.

# Definitions
- missing: The hypothesis fails to fully cover the reference.
  - Missing any entire product, or
  - Missing any required feature of any product, or
  - Only giving generic statements that do not confirm required features.
- wrong: The hypothesis introduces information not present in the reference.
  - Adds external products not in the reference, or
  - Claims features that are not listed in the reference, or
  - States information that contradicts the reference.

# Rules
- Treat matching semantically: synonyms and paraphrases count if they clearly refer to the same product/feature.
- Every product and its required features must be present.
- Ignore details not specified in the reference (e.g., price, seller, shipping, warranty) unless explicitly included as required features.
- If both missing and wrong apply, output both labels.
- Do not include explanations, reasoning, or any extra text.
- Only output the feature names, do not include the feature values.

# Reference
{reference}

# Hypothesis
{hypothesis}

# Output
Output a valid JSON object with "missing" and "wrong" fields.
- "missing": a valid JSON array of strings, each string is a missing feature name.
- "wrong": a valid JSON array of strings, each string is a wrong feature name.
- If the hypothesis exactly matches the reference, the "missing" and "wrong" fields should be empty arrays.
Example format:
```json
{{
    "missing": ["feature_name1", "feature_name2", ...],
    "wrong": ["feature_name3", "feature_name4", ...]
}}
```
\end{lstlisting}
\end{tcolorbox}
\caption{User Simulator Prompt for High Hint}
\label{fig:user_high_hint}
\end{figure*}

\section{Supplemental Details For Reward Design}
\subsection{Outcome-Level Reward}
\label{app:reward_prompts}

This appendix documents the LLM-as-judge protocol used to compute the stage rewards in Sec.~\ref{sec:shopping_reward_design}.
We employ four task-and-stage specific prompts, corresponding to two tasks (single-product vs.\ add-on-deals) and two stages (Stage-1 preference grounding vs.\ Stage-2 product matching).
All prompts are instantiated with (i) the user instruction, (ii) the agent output to be evaluated, and (iii) reference information from our benchmark annotations and product index.
The judge is required to return structured outputs in a strict JSON schema to enable deterministic parsing and reward computation.

\subsubsection{Stage-1: Preference Grounding Reward Prompt}
\label{app:reward_prompts_stage1}

Stage-1 rewards evaluate whether the agent correctly grounds user preferences from long-term conversations before product retrieval.
Given the user query and the agent's intermediate response, the judge assesses (i) query relevance and (ii) preference extraction quality measured by how many annotated preference attributes are surfaced.
For add-on-deals, the judge additionally evaluates whether the response identifies a correct number of products aligned with the reference bundle.

\subsubsection{Stage-2: Product Matching Reward Prompt}
\label{app:reward_prompts_stage2}

Stage-2 rewards evaluate whether the agent's recommended product(s) match the user intent and satisfy preference attributes, grounded in the provided product attributes/options.
Given the user query, wanted features, and the retrieved product descriptions, the judge assesses (i) query-intent relevance and (ii) preference satisfaction based on feature matches.
For add-on-deals, relevance is counted at the product level and feature matches are aggregated across the bundle.

\subsection{Tool-Wise Reward}
\label{app:reward_implementation}

The per-tool reward formulations are presented in Sec.~\ref{sec:shopping_reward_design}. During RL training, we implement a reward server that maintains indexed product and memory databases. After the agent executes a trajectory $\tau$, the server: (1) extracts all tool invocations and their arguments, (2) computes $r(u)$ for each tool call against gold-standard annotations, (3) aggregates via Eq.~\ref{eq:toolwise_mean} to obtain $R_{\mathrm{tool}}(\tau)$, and (4) combines with $R_z(\tau_z)$ and $R_{\mathrm{fmt}}(\tau)$ via Eq.~\ref{eq:final_reward}. This enables immediate per-tool feedback without waiting for terminal-state evaluation, significantly improving credit assignment and training efficiency.

\section{Experimental Implementation}
\label{appendix:Experimental Implementation}
\subsection{Dataset Details}
\label{appendix:dataset_details}

For supervised fine-tuning (SFT)\cite{ouyang2022training}, we generate successful trajectories via GPT-4.1 rejection sampling, yielding 2,948 step-level examples used to initialize the model.

For reinforcement learning, since our framework operates in two stages with distinct system prompts and tool sets (Stage-1 for preference identification and Stage-2 for shopping assistance), each instruction generates two separate data instances—one for each stage. This results in 1,600 training instances (800 queries $\times$ 2 stages) and 400 test instances (200 queries $\times$ 2 stages), totaling 2,000 instances. We train for 5 epochs with a learning rate of $1 \times 10^{-6}$.

\subsection{Training Details}
\paragraph{SFT.} For SFT, we employ LLaMA-Factory~\cite{zheng2024llamafactory}
and 8$\times$H20 GPUs to fine-tune Qwen3-4B-Thinking-2507 with LoRA (rank $= 64$)~\citep{hu2022lora},
targeting the query, key, value, and output projection layers
(\texttt{q\_proj}, \texttt{k\_proj}, \texttt{v\_proj}, \texttt{o\_proj}).
Training is conducted for 3 epochs with a cosine learning rate schedule
(peak learning rate $= 5 \times 10^{-5}$), using
BF16 mixed precision with an effective batch size of
4 per device.

\paragraph{RL.} For reinforcement learning, we utilize the VeRL framework\cite{sheng2024hybridflow} with GRPO algorithm\cite{shao2024deepseekmath}. Key RL hyperparameters include: eight rollouts per sample ($n=8$), maximum output length of 32,768 tokens, maximum 20 assistant turns per trajectory, batch size of 16, mini-batch size of 8, temperature of 0.6, top-$k$ sampling with $k=20$ and top-$p$ with $p=0.95$. We train for 2.6 epochs with a learning rate of $1 \times 10^{-6}$. The training is conducted on 8 NVIDIA H20 GPUs using FSDP for distributed training.

\section{Evaluation Robustness}
\label{sec:eval_robustness}

\paragraph{Multi-LLM Judge Consistency.}
To verify that our evaluation is not biased by a single judge model, we expand the LLM-as-Judge evaluation to five diverse models: GPT-5, Qwen3-Max, Kimi-K2, Gemini-3-Pro, and Qwen3-Next-80B. Table~\ref{tab:multi_judge} shows the success rates assigned by each judge to four agent systems.

\begin{table}[h]
\centering
\small
\setlength{\tabcolsep}{3pt}
\renewcommand{\arraystretch}{1.0}
\resizebox{\columnwidth}{!}{%
\begin{tabular}{lcccc}
\toprule
\textbf{Judge} & \textbf{GPT-5} & \textbf{GPT-4.1} & \textbf{GPT-4o} & \textbf{Qwen3-Max} \\
\midrule
GPT-5          & 64.5 & 51.0 & 49.0 & 48.0 \\
Qwen3-Max      & 59.0 & 52.0 & 43.0 & 45.5 \\
Kimi-K2        & 58.5 & 48.5 & 44.0 & 49.0 \\
Gemini-3-Pro   & 66.5 & 54.5 & 45.0 & 48.5 \\
Qwen3-Next-80B & 63.0 & 49.5 & 43.0 & 46.0 \\
\bottomrule
\end{tabular}
}
\caption{Success rates (\%) across different judge LLMs (columns are agent systems). All five judges consistently rank GPT-5 Agent as the top system.}
\label{tab:multi_judge}
\end{table}

All five judges produce a consistent ranking, with GPT-5 Agent ranked first by every judge. We compute the Overall Performance Correlation (OPC) via pairwise Pearson correlations among judges' score vectors, obtaining a mean of 0.9547 (std 0.0289), confirming that our conclusions are robust across diverse judge models.

\begin{figure*}[h]
\centering
\includegraphics[width=\textwidth]{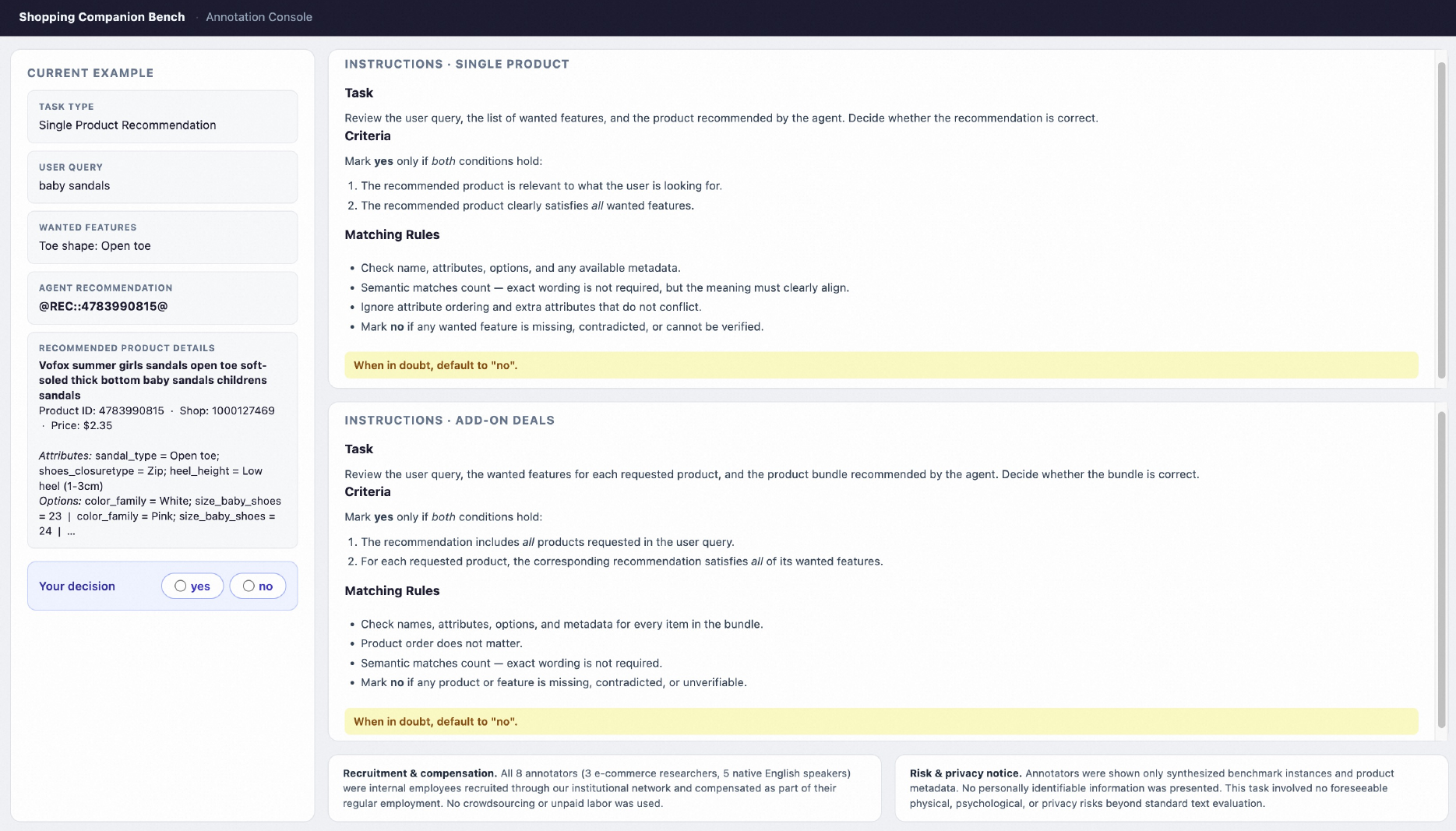}
\caption{Screenshot of the internal annotation console used for human meta-evaluation, showing the annotation instructions, an example instance, and the decision interface.}
\label{fig:annotation_console}
\end{figure*}

\paragraph{Human Meta-Evaluation.}
We conduct human meta-evaluation on 200 samples (100 per task), annotated by 8 internal annotators---3 researchers with over 5 years of e-commerce experience and 5 native English speakers---recruited through our institutional network and compensated as part of their regular employment. No external crowdsourcing or unpaid labor was used. Annotators were shown only synthesized benchmark instances and product metadata; no personally identifiable information was presented and the task involved no foreseeable risks beyond standard text evaluation. A screenshot of the annotation console with the full instructions given to annotators is shown in Figure~\ref{fig:annotation_console}.

Under a double-annotation protocol, we report \textbf{iPAR} (human inter-annotator agreement) and \textbf{PAR} (agreement between the GPT-5 judge and human consensus). As shown in Table~\ref{tab:human_meta_evaluation}, the results, iPAR = 0.8950 and PAR = 0.9497, show that human judgments are reliable and that the GPT-5 judge is highly aligned with human consensus.

\begin{table}[h]
\centering
\small
\setlength{\tabcolsep}{5pt}
\renewcommand{\arraystretch}{1.0}
\begin{tabular}{lccc}
\toprule
\textbf{Metric} & \textbf{Single} & \textbf{Add-on} & \textbf{Avg.} \\
\midrule
iPAR & 0.9200 & 0.8700 & 0.8950 \\
PAR  & 0.9674 & 0.9310 & 0.9497 \\
\bottomrule
\end{tabular}
\caption{Human meta-evaluation results. iPAR measures human inter-annotator agreement, while PAR measures agreement between the GPT-5 judge and human consensus.}
\label{tab:human_meta_evaluation}
\end{table}

\section{Preliminary Analysis of GPT-5 Zero-Shot Agent}
\label{appendix:error_analysis}


Despite strong general capabilities, GPT-5 achieves only 64.5\% success rate on our benchmark under zero-shot evaluation. To understand the nature of these failures, we first characterize the agent's behavioral patterns across all 200 test trajectories, and then systematically categorize the 69 failed cases.

\subsection{Agent Behavioral Patterns}
\label{appendix:behavior_patterns}

Figure~\ref{fig:behavior_analysis} presents two complementary views of GPT-5 zero-shot agent behavior: the per-turn tool call distribution (left) and the failure mode breakdown (right).

\paragraph{Phase 1: Memory Retrieval (Turns 0--2).}
All 200 trajectories begin with \texttt{mem\_search} at Turn~0, reflecting a universal ``retrieve preferences first'' strategy. At Turn~1, \texttt{mem\_view} calls peak as the agent expands relevant dialogue sessions, while \texttt{product\_search} calls begin in parallel. By Turn~2, memory tools recede and product tools dominate---indicating that the agent typically completes preference extraction within 2--3 turns.

\paragraph{Phase 2: Product Search--View Loop (Turns 3+).}
From Turn~3 onward, the agent enters a \texttt{product\_search} and \texttt{product\_view} loop that dominates the remainder of the trajectory. Frequency peaks at Turns~3--4 (approximately 490 total tool calls per turn across 200 trajectories) and gradually decays as simpler tasks complete. The long tail beyond Turn~12 corresponds to complex add-on deals requiring multiple products from the same shop, where the agent must iterate through numerous candidates.

\paragraph{Early Errors Cascade.}
A critical observation is that errors in Phase~1 propagate through Phase~2. When the agent extracts incorrect or incomplete preferences from memory (Preference Hallucination, 10.1\%), all subsequent product searches are misguided. Similarly, when the agent skips \texttt{mem\_view} entirely and relies on snippet-level \texttt{mem\_search} results (Insufficient Tool Calls, 10.1\%), it lacks the context needed for accurate preference identification. These early-stage errors account for 20.2\% of all failures and are particularly costly because they cannot be recovered through better product search strategies.

\paragraph{Verification Gaps in Phase 2.}
Even with correct preferences, the agent frequently fails during product verification. Unverified Attributes (39.1\%)---the dominant error type---occurs when the agent infers product features from titles alone without calling \texttt{product\_view} to confirm specifications. This pattern is visible in the relatively low proportion of \texttt{product\_view} calls compared to \texttt{product\_search}: the agent searches aggressively but verifies insufficiently.

\subsection{Error Taxonomy}
\label{appendix:error_taxonomy}

We analyze the 69 failed trajectories using a two-stage LLM-based attribution pipeline: first, free-form per-trajectory analysis of each tool call and its consequences; then, bottom-up clustering into unified error categories. Table~\ref{tab:error_distribution} summarizes the resulting distribution.

\begin{table}[h]
\centering
\small
\setlength{\tabcolsep}{4pt}
\begin{tabular}{l c c c}
\toprule
\textbf{Error Type} & \textbf{Single} & \textbf{Add-on} & \textbf{Total} \\
\midrule
Unverified Attributes & 10 & 17 & 27 (39.1\%) \\
Same-shop Alignment & --- & 11 & 11 (15.9\%) \\
Multi-SKU in Single Task & 9 & --- & 9 (13.0\%) \\
Product ID / Sourcing & 3 & 5 & 8 (11.6\%) \\
Preference Hallucination & 2 & 5 & 7 (10.1\%) \\
Insufficient Tool Calls & --- & 7 & 7 (10.1\%) \\
\midrule
\textbf{Total} & \textbf{24} & \textbf{45} & \textbf{69} \\
\bottomrule
\end{tabular}
\caption{Distribution of error types across 69 failed GPT-5 zero-shot trajectories.}
\label{tab:error_distribution}
\vspace{-1.5em}
\end{table}

\paragraph{Error Categories.} We identify six categories, each corresponding to a distinct failure mode in tool-augmented shopping trajectories:

\begin{itemize}[leftmargin=*, topsep=2pt, itemsep=1pt, parsep=0pt]
\item \textbf{Unverified Attributes:} The agent asserts product features (e.g., ``ultrasonic'', ``pet-safe'') not confirmed by \texttt{product\_view}, typically inferred from product titles alone.
\item \textbf{Same-shop Alignment:} The agent combines products across shops without checking Shop~ID consistency, violating voucher requirements.
\item \textbf{Multi-SKU in Single Task:} The agent outputs multiple product IDs for a task that requires exactly one recommendation---an instruction-following failure.
\item \textbf{Product ID / Sourcing Errors:} The agent fabricates or misattributes specifications and prices with no provenance in tool responses.
\item \textbf{Preference Hallucination:} The agent claims user preferences without verification through \texttt{mem\_view}, relying on snippet-level \texttt{mem\_search} fragments.

\item \textbf{Insufficient Tool Calls:} The agent skips necessary tool invocations entirely, outputting placeholder recommendations without evidence.
\end{itemize}

\begin{figure*}[h]
\centering
\begin{tcolorbox}[
    enhanced,
    colback=red!2!white,
    colframe=red!50!black,
    fonttitle=\bfseries\sffamily,
    title={\small Case Study 1: Unverified Attributes},
    coltitle=white,
    attach boxed title to top left={yshift=-2mm, xshift=3mm},
    boxed title style={colback=red!50!black, sharp corners},
    sharp corners,
    boxrule=0.5pt,
    left=3pt, right=3pt, top=8pt, bottom=3pt, arc=0pt
]
\small
\textbf{Task:} Single Product \quad \textbf{Query:} \textit{``lizard repellent''}

\medskip
\textbf{User Context:} Ground-floor apartment; small lizards at night; has a toddler and a curious cat; wants eco-friendly, odorless, set-and-forget solution.

\medskip
\textbf{Key Trajectory:}

\textit{Step 3} --- \texttt{product\_search("eco-friendly lizard repellent pet safe indoor low odor placement tablets sachets granules")} $\rightarrow$ 50 results returned.

\textit{Step 6} --- \texttt{product\_view({[}5235454373{]})} $\rightarrow$ Returns:

\smallskip
\hspace{1em}{\ttfamily\scriptsize Attributes: Features = With Lights, Portable, Eco-Friendly; Options: (empty)}
\smallskip

\textit{Step 7 (Final Answer)} --- Agent claims: \textit{``Ultrasonic plug-in pest repellent; 100\% safe for humans and pets; electronic ultrasound---no spray, no odor.''}

\medskip
\textbf{\color{red!70!black} Error:} The \texttt{product\_view} response contains \textbf{no} ``ultrasonic'', ``pet-safe'', or ``lizard'' attributes. The agent inferred these from the product \emph{title} in search results without verification, asserting critical safety and functionality claims unsupported by structured attributes.
\end{tcolorbox}
\caption{Case Study: Unverified Attributes. The agent asserts ``ultrasonic'' and ``pet-safe'' features that do not appear in the \texttt{product\_view} response.}
\label{fig:case_unverified_attrs}
\end{figure*}

\begin{figure*}[h]
\centering
\begin{tcolorbox}[
    enhanced,
    colback=orange!3!white,
    colframe=orange!60!black,
    fonttitle=\bfseries\sffamily,
    title={\small Case Study 2: Same-Shop Alignment Failure},
    coltitle=white,
    attach boxed title to top left={yshift=-2mm, xshift=3mm},
    boxed title style={colback=orange!60!black, sharp corners},
    sharp corners,
    boxrule=0.5pt,
    left=3pt, right=3pt, top=8pt, bottom=3pt, arc=0pt
]
\small
\textbf{Task:} Add-on Deals \quad \textbf{Query:} \textit{``Bundle: notebook, cereal, and whiskey. Voucher: valid when total > \$22, 50\% off up to \$21. Budget: \$21.''}

\medskip
\textbf{Key Trajectory:}

\textit{Step 4} --- Three parallel \texttt{product\_search} calls return:

\smallskip
\hspace{1em}{\ttfamily\scriptsize Jack Daniels 700ml -- ID: 991722797, \textbf{Shop: 8921}, \$21.20}\\
\hspace{1em}{\ttfamily\scriptsize Energen Cereal 40g{$\times$}30 -- ID: 412486028, \textbf{Shop: 1000229478}, \$5.02}\\
\hspace{1em}{\ttfamily\scriptsize A5 Notebook 180pp -- ID: 4448768196, \textbf{Shop: 500956320146}, \$2.97}
\smallskip

\textit{Step 9 (Final Answer)} --- Agent combines all three items, computes voucher discount (\$29.19 $\rightarrow$ \$14.60), and recommends the bundle.

\medskip
\textbf{\color{orange!70!black} Error:} The three products come from \textbf{three different shops} (8921, 1000229478, 500956320146). The agent never checked or constrained selections to a single merchant, making the voucher inapplicable in practice. Shop ID information was clearly visible in every \texttt{product\_search} response but entirely ignored.
\end{tcolorbox}
\caption{Case Study: Same-Shop Alignment Failure. The agent assembles products from three different shops but applies a same-shop voucher without verifying merchant consistency.}
\label{fig:case_same_shop}
\end{figure*}

\begin{figure*}[h]
\centering
\begin{tcolorbox}[
    enhanced,
    colback=yellow!5!white,
    colframe=yellow!50!black,
    fonttitle=\bfseries\sffamily,
    title={\small Case Study 3: Multi-SKU in Single Task},
    coltitle=white,
    attach boxed title to top left={yshift=-2mm, xshift=3mm},
    boxed title style={colback=yellow!50!black, sharp corners},
    sharp corners,
    boxrule=0.5pt,
    left=3pt, right=3pt, top=8pt, bottom=3pt, arc=0pt
]
\small
\textbf{Task:} Single Product \quad \textbf{Query:} \textit{``USB Type-C cable''}

\medskip
\textbf{User Context:} Splits time between home/office; USB-C phone and laptop; wants dependable cable; discussed 65W charger inclusion in past conversation.

\medskip
\textbf{Key Trajectory:}

\textit{Step 4} --- Agent searches: \texttt{product\_search("USB-C to USB-C cable 65W charger bundle")} --- search strategy already biased toward bundles.

\textit{Steps 5--6} --- \texttt{product\_view} on separate charger and cable products (UGREEN GaN charger, various C-to-C cables).

\textit{Step 7 (Final Answer)} --- Agent recommends: \textit{``The bundle below pairs a reputable 65W GaN charger with a USB-C to USB-C cable''} --- outputting \textbf{two separate product IDs}.

\medskip
\textbf{\color{yellow!50!black} Error:} The task type is \texttt{single\_product}, requiring exactly \textbf{one} product ID. The agent assembled two independent SKUs (charger + cable) instead of finding a single listing that includes both, or simply recommending only the requested cable.
\end{tcolorbox}
\caption{Case Study: Multi-SKU in Single Task. The agent outputs two product IDs for a task that requires exactly one.}
\label{fig:case_multi_sku}
\end{figure*}

\begin{figure*}[h]
\centering
\begin{tcolorbox}[
    enhanced,
    colback=purple!3!white,
    colframe=purple!50!black,
    fonttitle=\bfseries\sffamily,
    title={\small Case Study 4: Preference Hallucination},
    coltitle=white,
    attach boxed title to top left={yshift=-2mm, xshift=3mm},
    boxed title style={colback=purple!50!black, sharp corners},
    sharp corners,
    boxrule=0.5pt,
    left=3pt, right=3pt, top=8pt, bottom=3pt, arc=0pt
]
\small
\textbf{Task:} Add-on Deals \quad \textbf{Query:} \textit{``Bundle: baby shoes, network connectors, and slime glitter. Voucher: valid on totals > \$7, 30\% off capped at \$3. Budget: \$9.''}

\medskip
\textbf{Key Trajectory:}

\textit{Step 1} --- \texttt{mem\_search({[}20+ queries{]})} $\rightarrow$ Returns snippet-level results:

\smallskip
\hspace{1em}{\ttfamily\scriptsize {[}Index 96{]} user: Hook and Loop would be wanted...}\\
\hspace{1em}{\ttfamily\scriptsize {[}Index 377{]} assistant: The brand on these is ZOERAX...}\\
\hspace{1em}{\ttfamily\scriptsize {[}Index 461{]} user: I want Neon Pink and Glitter or Sequins Added...}
\smallskip

\textit{Step 2 (Stage 1 Answer)} --- Agent outputs detailed preferences: \textit{``Hook-and-loop strap is wanted; ZOERAX is preferred; 30 pieces wanted; Neon Pink wanted''} --- \textbf{without ever calling \texttt{mem\_view}}.

\medskip
\textbf{\color{purple!70!black} Error:} \texttt{mem\_search} returns only single-sentence snippets without surrounding context. The agent skipped \texttt{mem\_view} (which retrieves full dialogue sessions) and directly treated fragments as confirmed preferences. These snippets may represent options \emph{discussed} but not finalized, leading to over-constrained searches and mismatched recommendations downstream.
\end{tcolorbox}
\caption{Case Study: Preference Hallucination. The agent skips \texttt{mem\_view} and treats \texttt{mem\_search} snippets as confirmed preferences without verifying full dialogue context.}
\label{fig:case_pref_hallucination}
\end{figure*}

\begin{figure*}[h]
\centering
\begin{tcolorbox}[
    enhanced,
    colback=blue!3!white,
    colframe=blue!50!black,
    fonttitle=\bfseries\sffamily,
    title={\small Case Study 5: Product ID / Sourcing Errors},
    coltitle=white,
    attach boxed title to top left={yshift=-2mm, xshift=3mm},
    boxed title style={colback=blue!50!black, sharp corners},
    sharp corners,
    boxrule=0.5pt,
    left=3pt, right=3pt, top=8pt, bottom=3pt, arc=0pt
]
\small
\textbf{Task:} Add-on Deals \quad \textbf{Query:} \textit{``Bundle: solar charge controller, WiFi dongle, and solar inverter. Same-shop voucher: total > \$453, 30\% off capped at \$338. Budget: \$405.''}

\medskip
\textbf{Key Trajectory:}

\textit{Step 8} --- \texttt{product\_view({[}4591396209, 4994321724, ...{]})} $\rightarrow$ Returns:

\smallskip
\hspace{1em}{\ttfamily\scriptsize Product 4591396209: Ampere\_Rating=80; Type\_Of\_Electrical\_Circuit\_Parts=Load;}\\
\hspace{1em}{\ttfamily\scriptsize \phantom{Product 4591396209: }brand=Flowatt; material=Stainless steel}\\
\hspace{1em}{\ttfamily\scriptsize Product 4994321724: Type\_Of\_Electrical\_Connectors=Wire/Cable-To-Board;}\\
\hspace{1em}{\ttfamily\scriptsize \phantom{Product 4994321724: }brand=Flowatt}
\smallskip

\textit{Step 11 (Final Answer)} --- Agent claims: \textit{``Flowatt 3 kW hybrid off-grid inverter, 24V battery input, \$160.94''} and \textit{``WiFi Module with RS232, \$2.44''}.

\medskip
\textbf{\color{blue!70!black} Error:} The \texttt{product\_view} attributes contain \textbf{no} voltage, wattage, inverter type, WiFi capability, or price information. The agent fabricated specifications (24V, 3kW, RS232) from product \emph{titles} seen in search results and presented them as ``verified'' attributes. Prices quoted (\$160.94, \$2.44) also have no tool-response provenance.
\end{tcolorbox}
\caption{Case Study: Product ID / Sourcing Errors. The agent claims verified specifications (24V, 3kW, RS232, prices) that are absent from \texttt{product\_view} responses.}
\label{fig:case_product_id_errors}
\end{figure*}

\begin{figure*}[h]
\centering
\begin{tcolorbox}[
    enhanced,
    colback=teal!3!white,
    colframe=teal!50!black,
    fonttitle=\bfseries\sffamily,
    title={\small Case Study 6: Insufficient Tool Calls},
    coltitle=white,
    attach boxed title to top left={yshift=-2mm, xshift=3mm},
    boxed title style={colback=teal!50!black, sharp corners},
    sharp corners,
    boxrule=0.5pt,
    left=3pt, right=3pt, top=8pt, bottom=3pt, arc=0pt
]
\small
\textbf{Task:} Add-on Deals \quad \textbf{Query:} \textit{``Bundle: breast pump and washing machine cleaner. Voucher: valid when total > \$15, 20\% off capped at \$7. Budget: \$19.''}

\medskip
\textbf{Key Trajectory (3 steps total):}

\textit{Step 1} --- \texttt{mem\_search({[}``breast pump'', ``washing machine cleaner'', ...{]})} $\rightarrow$ Returns user preferences (double electric pump, HE-compatible cleaner).

\textit{Step 2} --- Agent outputs Stage 1 preference summary.

\textit{Step 3 (Final Answer)} --- Agent directly outputs a recommendation with budget math (\textit{``pre-discount bundle: \$15.01--\$23.75''}) but \textbf{never calls \texttt{product\_search} or \texttt{product\_view}}. Product IDs in the output are placeholders with no tool-response backing.

\medskip
\textbf{\color{teal!70!black} Error:} The entire Stage 2 (shopping execution) contains \textbf{zero tool calls}. The agent performed mental arithmetic on voucher thresholds and fabricated a ``recommendation'' without querying any product data. This represents the most extreme form of tool underutilization---the agent simulated task completion without grounding in any external evidence.
\end{tcolorbox}
\caption{Case Study: Insufficient Tool Calls. The agent completes Stage 2 with zero \texttt{product\_search} or \texttt{product\_view} calls, outputting fabricated product IDs.}
\label{fig:case_insufficient_tools}
\end{figure*}

\end{document}